\documentclass[runningheads]{llncs}

\usepackage{eccv}

\usepackage{eccvabbrv}
\usepackage{marvosym}
\usepackage{graphicx}
\usepackage{booktabs}
\usepackage{fontawesome}
\usepackage{amsmath}
\usepackage{mathtools}
\usepackage{multirow}
\usepackage{multicol}
\usepackage{makecell}
\usepackage{xcolor}
\usepackage{colortbl}
\usepackage{makecell}
\usepackage{pifont}

\usepackage[accsupp]{axessibility}  % Improves PDF readability for those with disabilities.

\newcommand{\mathtiny}[1]{%
  \mathchoice
    {\scalebox{0.7}{$\displaystyle #1$}}% display
    {\scalebox{0.7}{$\textstyle #1$}}%   text
    {\scalebox{0.7}{$\scriptstyle #1$}}% script
    {\scalebox{0.7}{$\scriptscriptstyle #1$}}% scriptscript
}

\newcommand{\mathttt}[1]{%
  \mathchoice
    {\scalebox{0.6}{$\displaystyle #1$}}% display
    {\scalebox{0.6}{$\textstyle #1$}}%   text
    {\scalebox{0.6}{$\scriptstyle #1$}}% script
    {\scalebox{0.6}{$\scriptscriptstyle #1$}}% scriptscript
}

\newcommand{\xmark}{{\color[HTML]{000000}\ding{55}}}   
\newcommand{\cmark}{{\color[HTML]{000000}\ding{51}}} 

\definecolor{myred}{HTML}{C00000}
\definecolor{mygreen}{HTML}{70AD47}

\usepackage{hyperref}

\usepackage{orcidlink}

\begin{document}

% ---------------------------------------------------------------
% TODO REVIEW: Replace with your title
\title{LogiCo: A Unified Framework for Logical and Structural Anomaly Detection} 

% TODO REVIEW: If the paper title is too long for the running head, you can set
% an abbreviated paper title here. If not, comment out.
\titlerunning{LogiCo: A Unified Framework for Logical and Structural Anomaly Detection}

% TODO FINAL: Replace with your author list. 
% Include the authors' OCRID for the camera-ready version, if at all possible.
\author{Ximiao Zhang\inst{1} \and
Min Xu\inst{2} \and
Xiuzhuang Zhou\inst{1}\textsuperscript{\Letter}}

% TODO FINAL: Replace with an abbreviated list of authors.
\authorrunning{X. Zhang et al.}
% First names are abbreviated in the running head.
% If there are more than two authors, 'et al.' is used.

% TODO FINAL: Replace with your institution list.
\institute{School of Intelligent Engineering and Automation, Beijing University of Posts and Telecommunications, Beijing, China \\ \email{xiuzhuang.zhou@bupt.edu.cn}\and
College of Information and Engineering, Capital Normal University, Beijing, China}

\maketitle

\begin{abstract}
  Current anomaly detection methods primarily focus on structural anomalies, while paying insufficient attention to anomalies that violate logical constraints. Conversely, top-performing logical anomaly detection approaches address this by modeling global semantic consistency, but perform poorly on subtle structural anomalies due to inadequate detection granularity. In this paper, we propose \textbf{LogiCo}, a unified framework for \textbf{Logi}cal and structural anomaly detection via \textbf{Co}mponent-level feature reconstruction. Unlike existing methods that rely on explicit global semantic modeling, LogiCo employs a novel component-level feature reconstruction technique to capture inter-component logical constraints. Specifically, LogiCo maps pre-trained image features into a discrete component-level feature space and performs collaborative feature reconstruction at both component and patch levels, enabling it to effectively detect both logical and structural anomalies. Furthermore, to address the specific challenge of count-related logical anomalies, we integrate a segmentation-map discriminator that extends the model's capability to identify quantitative inconsistencies. LogiCo achieves state-of-the-art performance on both logical and structural anomaly detection across four benchmarks, including MVTec-LOCO, MVTec-AD, VisA, and Real-IAD, demonstrating its superiority and practical feasibility. The code is available at \url{https://github.com/cnulab/LogiCo}.
\end{abstract}

\section{Introduction}
\label{sec:intro}

\begin{figure}[t]
  \centering
   \includegraphics[width=\linewidth]{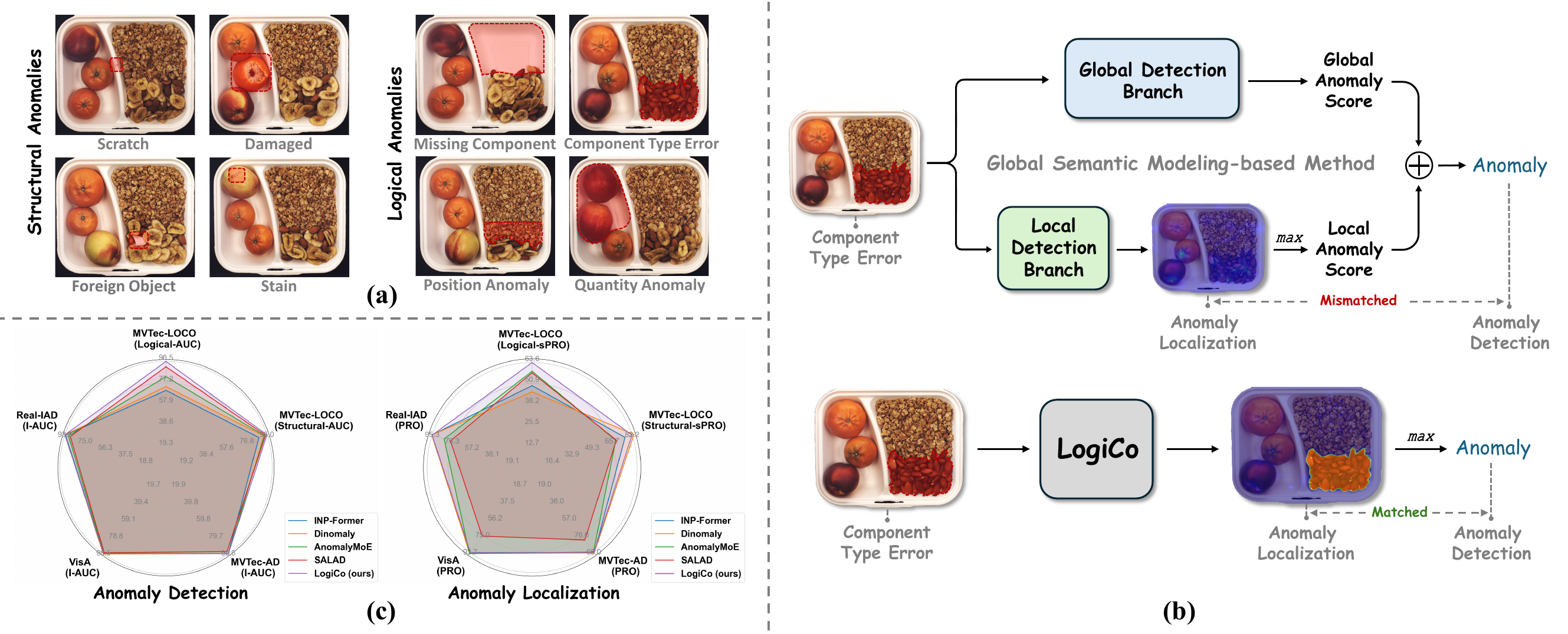}
   \caption{(a): Examples of structural and logical anomalies. (b): Comparison of the overall pipelines between LogiCo and global semantic modeling-based methods. (c): Comparison of LogiCo with other methods on anomaly detection and localization performance.}
   \label{fig:fig1}
\end{figure}

Image anomaly detection has emerged as a fundamental problem in industrial vision. Given the inherent scarcity of defective samples in real-world scenarios, this task is typically formulated as an unsupervised learning paradigm \cite{tao2022deep, diers2023survey, liu2024deep}. Anomalies are generally classified into two types: structural anomalies and logical anomalies. Structural anomalies are characterized by unseen local visual patterns, such as scratches, breakage, and stains. In contrast, logical anomalies represent invalid configurations that deviate from production flows or assembly requirements, such as missing components, incorrect locations, and count anomalies, as illustrated in \cref{fig:fig1}(a). Prevalent unsupervised anomaly detection approaches \cite{roth2022towards, defard2021padim, deng2022anomaly, zhang2024realnet, guo2025dinomaly, zhang2025towards} typically model the feature distribution of image patches. While effective for structural defects, these methods are intrinsically constrained by their local modeling schemes, failing to capture long-range logical dependencies. To mitigate this, recent state-of-the-art (SOTA) logical anomaly detection methods \cite{Hsieh_2024_BMVC, fuvcka2025salad, sugawara2024puad, kim2024few, peng2026vllm} usually employ an additional global detection branch to learn the global semantic consistency, forming a global-local architecture, as shown in \cref{fig:fig1}(b). The global branch discards spatial structure, allowing it to effectively capture overall appearance patterns and thereby help identify logical anomalies. Nevertheless, due to the lack of fine-grained discriminative information, the global branch struggles to detect subtle structural anomalies, which may impair the model's capability for structural anomaly detection. Furthermore, the global branch is typically unable to predict pixel-level anomaly maps, leading to mismatched anomaly detection and localization results. Currently, the performance of logical anomaly detection methods in detecting structural anomalies is still notably inferior to that of specialized structural detection methods.

To address the above issues, we propose LogiCo, a unified framework for logical and structural anomaly detection. Unlike previous works limited by coarse-grained global consistency modeling, LogiCo consistently preserves the image's spatial layout, thereby facilitating precise anomaly detection and localization, as illustrated in \cref{fig:fig1}(b). LogiCo consists of three key components: a Component-level Reconstruction Network (CRN), a Structural Reconstruction Network (SRN), and a Segmentation Map Discriminator (SMD). Specifically, the CRN is trained to restore discretized component-level feature maps to their normal pre-trained counterparts, thereby capturing inter-component logical inconsistencies. The SRN reconstructs fine-grained local features under cross-attention guidance, enabling precise detection of structural anomalies. The SMD models quantity constraints by learning the normal patterns of component segmentation maps, facilitating the identification of count-related anomalies. Compared with existing methods, LogiCo significantly extends the scope of detectable anomalies while preserving high detection accuracy. As illustrated in \cref{fig:fig1}(c), LogiCo surpasses existing methods in logical anomaly detection, achieving SOTA performance in both anomaly detection and localization. For structural anomalies, it achieves performance comparable to or exceeding that of top-tier methods. Our main contributions are summarized as follows:

\begin{itemize}
\item We propose LogiCo, a unified framework for logical and structural anomaly detection. LogiCo captures local structural consistency and global logical constraints in a fine-grained discriminative space, effectively addressing the limitations of existing methods regarding detection scope and granularity.
\item We introduce a novel component-level feature reconstruction technique, which transforms the complex modeling of multiple logical constraints into a unified reconstruction paradigm. This allows for the precise detection and pixel-level localization of logical anomalies without compromising spatial structures.
\item LogiCo achieves SOTA logical and structural anomaly detection performance on four benchmarks: MVTec-LOCO, MVTec-AD, VisA, and Real-IAD, demonstrating its effectiveness and potential for real-world applications.
\end{itemize}

\section{Related Work}
\label{sec:related}

\textbf{Structural Anomaly Detection.} Structural anomalies are prevalent in fields such as industrial manufacturing \cite{huang2020surface, bergmann2019mvtec, bergmann2021mvtec, zou2022spot, wang2024real} and medical imaging \cite{wyatt2022anoddpm, cai2022dual, xiang2023squid, zhang2024mediclip}, prompting extensive research. Since such anomalies typically manifest as localized abnormal patterns, existing methods usually employ local feature modeling schemes. Among them, reconstruction-based methods \cite{zavrtanik2021draem, zavrtanik2022dsr, you2022unified, zhang2023exploring, fuvcka2024transfusion} train models to learn the normal patterns of images, and identify anomalous regions via pixel-wise or patch-wise reconstruction. Knowledge distillation-based methods \cite{bergmann2020uninformed, deng2022anomaly, tien2023revisiting, batzner2024efficientad, guo2025dinomaly} train a student network to learn the normal feature distribution of the teacher network. In the inference phase, they localize anomalous regions via patch-level feature comparison. Deep feature embedding-based methods learn the normal feature distribution during training, for instance, by constructing memory banks \cite{roth2022towards,bae2023pni,damm2025anomalydino} or utilizing normalizing flows \cite{gudovskiy2022cflow, yu2021fastflow, lei2023pyramidflow}. Subsequently, anomaly detection is achieved by measuring the deviation between query features and the learned feature distribution. Despite these advances, the above methods lack global perception capabilities, and thus often fail in detecting logical anomalies. Anomaly synthesis-based methods \cite{li2021cutpaste, schluter2022natural, zhang2024realnet, hu2024anomalydiffusion} train anomaly detection models using synthetic anomalous samples. These approaches emphasize that synthetic anomalies can effectively simulate real-world anomalies. However, for logical anomaly detection, synthesizing realistic logical anomalous samples in either the pixel space or the continuous feature space remains a significant challenge \cite{zhao2024logical, tong2025component}.

\textbf{Logical Anomaly Detection.} Logical anomaly detection focuses on capturing global semantic constraints and long-range dependencies; it has broad application prospects yet remains underexplored. Bergmann et al. \cite{bergmann2022beyond} released MVTec-LOCO, the first dataset for logical anomaly detection, and proposed GCAD, a logical anomaly detection framework based on global-local knowledge distillation. GCAD employs a global network branch with a low-dimensional bottleneck to learn global context and overall semantic constraints. EfficientAD \cite{batzner2024efficientad} inherits the core idea of GCAD and identifies logical anomalies by comparing the reconstruction error between two reconstruction networks with different model capacities. Sugawara et al. \cite{sugawara2024puad} observed that EfficientAD remains ineffective at detecting certain types of logical anomalies, such as missing components or count-related anomalies. To address this, they introduced a global detection branch that disregards inter-component spatial relationships and instead models global semantic consistency using a Gaussian probability model \cite{rippel2021modeling} to predict image-level anomaly scores. Recent works such as CSAD\cite{Hsieh_2024_BMVC}, SALAD\cite{fuvcka2025salad}, and PSAD\cite{kim2024few} have integrated global semantic modeling with component segmentation techniques, achieving SOTA performance in logical anomaly detection. However, due to the limited detection granularity of the global branch, these methods struggle with subtle structural anomalies, thereby limiting their practical application.

\section{Method}

\begin{figure}[t]
  \centering
   \includegraphics[width=\linewidth]{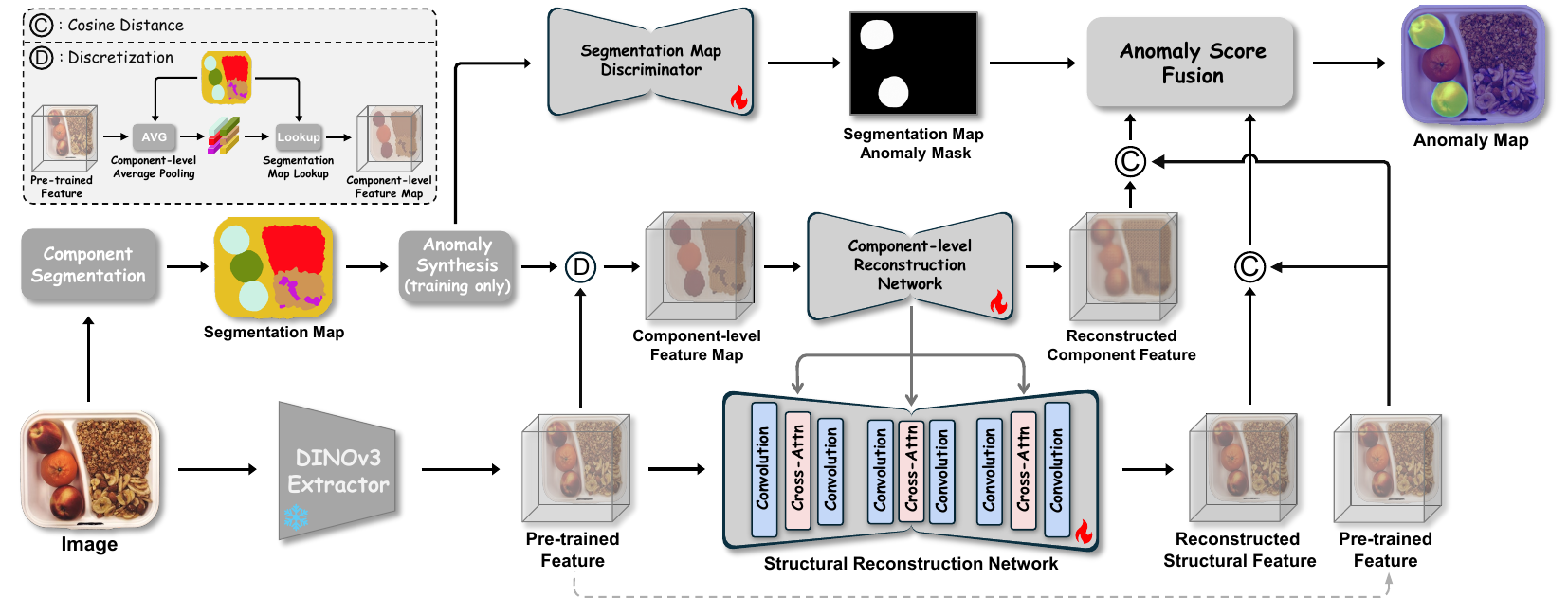}
   \caption{Overview of the proposed \textbf{LogiCo} pipeline, which consists of a Component-level Reconstruction Network (CRN), a Structural Reconstruction Network (SRN), and a Segmentation Map Discriminator (SMD). The CRN and SRN learn global logical constraints and local structural consistency to detect logical and structural anomalies, respectively, whereas the SMD learns quantity rules to identify count-related anomalies.}
   \label{fig:fig2}
\end{figure}

The overall pipeline of the proposed LogiCo is illustrated in \cref{fig:fig2}. Given an input image, LogiCo first employs open-vocabulary semantic segmentation to generate component segmentation maps. Guided by these maps, pre-trained image features are discretized and mapped into a component-level feature space. During the training phase, LogiCo synthesizes pseudo-anomalous features by manipulating the segmentation maps. A Component-level Reconstruction Network (CRN) is then trained to restore these features back to their normal, pre-trained counterparts, thereby capturing underlying logical inconsistencies. For structural anomalies, LogiCo employs cross-attention guided reconstruction to compel the Structural Reconstruction Network (SRN) to restore the anomaly features, ensuring comprehensive recall of the anomalous regions. Additionally, a Segmentation Map Discriminator (SMD) is integrated to identify irregularities in segmentation maps, enhancing the detection of quantity-related anomalies.

\subsection{Component-level Feature Reconstruction}

We employ DINOv3 \cite{simeoni2025dinov3} for feature extraction and open-vocabulary semantic segmentation, which aligns image and text embeddings into a joint embedding space. Specifically, for an input image $I \in \mathbb{R}^{H \times W \times 3}$, we define a set of text prompts $\{P_1, P_2, \dots, P_K\}$ to describe its component categories, such as \{``\textit{Peach}'', ``\textit{Orange}'',...\}, where $H$ and $W$ represent the height and width of the image, respectively, and $K$ is the number of component types. We then extract the pre-trained feature $ X=\phi_\mathtt{v}(I) \in \mathbb{R}^{h \times w \times c}$ and project it into the joint embedding space, yielding $ X_\mathtt{p}=\phi_\mathtt{p}(X) \in \mathbb{R}^{h \times w \times c_\mathtt{p}}$, where $\phi_\mathtt{v}(\cdot)$ and $\phi_\mathtt{p}(\cdot)$ denote the DINOv3 vision encoder and the projection head, respectively, while $c$ and $c_\mathtt{p}$ represent the feature dimensions before and after projection. For text prompts, we use the DINOv3 text encoder $\phi_\mathtt{t}(\cdot)$ to embed them into the embedding space to obtain $\{\mathcal{P}_1, \mathcal{P}_2, \dots, \mathcal{P}_K\}$, where $\mathcal{P}_k = \phi_\mathtt{t}(P_k) \in \mathbb{R}^{c_\mathtt{p}}$ denotes the embedding of the $k$-th prompt. Utilizing the shared embedding space between text embeddings and projected image features, we compute their similarity at each spatial location and then generate the segmentation map $G \in \{1,2,\dots,K\}^{h \times w}$ by assigning each patch to the component category with the highest similarity: 
\begin{equation}
\label{eq:equ1}
G(i,j)=\mathop{\arg\max}_{k \in \{1,2,\dots,K \}}\cos(X_\mathtt{p}(i,j),\mathcal{P}_k), 
\end{equation}
where $(i,j)$ denotes the spatial coordinates. Finally, we apply a simple post-processing procedure to refine the segmentation maps, for example by filtering out very small connected components. In contrast to recent works \cite{Hsieh_2024_BMVC, kim2024few, fuvcka2025salad, peng2025sam}, we do not rely on overly refined segmentation maps to avoid additional computational overhead. 

Based on the generated segmentation map, we transform the pre-trained feature into a discretized component-level representation $X_\mathtt{c} \in \mathbb{R}^{h \times w \times c}$. Specifically, we first compute the component prototypes $\{\mathrm{f}_1, \mathrm{f}_2, \cdots, \mathrm{f}_K\}$ via component-level average pooling:
\begin{equation}
\label{eq:equ2}
\mathrm{f}_k= \frac{1}{\lvert \Omega_k \rvert}\sum_{(i,j) \in \Omega_k}X(i,j),
\end{equation}
where $\Omega_k= \{(i,j)|G(i,j)=k\}$ denotes the spatial indices of the $k$-th component. Then, we construct the discretized feature map $X_\mathtt{c}$ via a lookup operation by assigning the corresponding component prototype to each spatial location: 
\begin{equation}
\label{eq:equ3}
X_\mathtt{c}(i,j) = \mathrm{f}_{G(i,j)}.
\end{equation}
It offers two key advantages: (1) The component-level feature map preserves the image's spatial layout and essential information for logical anomaly detection while filtering out irrelevant textural details, thereby facilitating the modeling of logical consistency. (2) The discretized feature space facilitates anomaly synthesis; specifically, by manipulating the segmentation maps during the lookup phase, we can generate diverse logical anomalies to enhance model training.

\begin{figure}[t]
  \centering
   \includegraphics[width=0.95\linewidth]{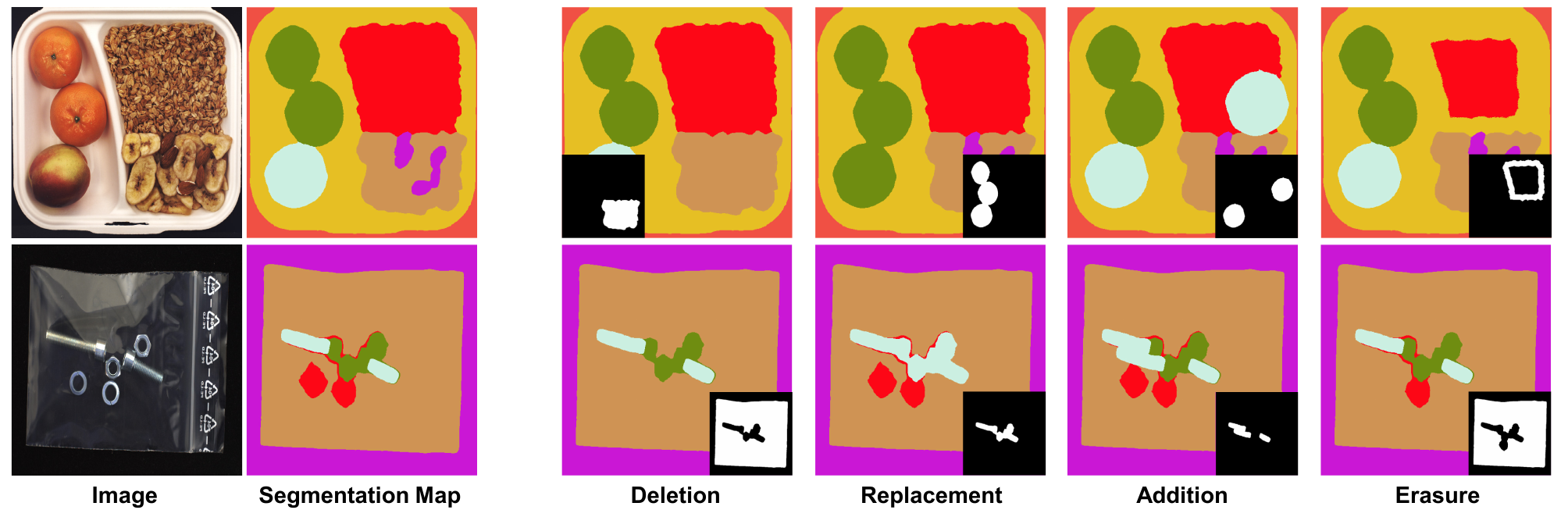}
   \caption{Examples of segmentation maps generated by various anomaly synthesis methods.}
   \label{fig:fig3}
\end{figure}

We design four anomaly synthesis strategies for LogiCo: component deletion, replacement, addition, and erasure, as shown in \cref{fig:fig3}. \textbf{Deletion}: Replacing the component with its neighboring background to remove it. \textbf{Replacement}: Replacing one component with another. \textbf{Addition}: Employing CutPaste \cite{li2021cutpaste} to duplicate a component at a different location, thereby increasing the component count. \textbf{Erasure}: Distinct from Deletion, Erasure only removes partial regions of a component. When a component consists of a single connected region, Erasure performs an erosion operation to alter its shape. When it comprises multiple connected regions, Erasure deletes parts of regions to reduce the component count. By integrating these methods, we generate diverse anomalous segmentation maps, which are then used to construct the anomalous component-level feature maps $X'_\mathtt{c}$. Subsequently, a Component-level Reconstruction Network (CRN) $\psi_{\mathtt{c}}$ is trained to restore these features into their normal pre-trained representations: $X_{\mathtt{L}} = \psi_{\mathtt{c}}(X'_\mathtt{c}) \in \mathbb{R}^{h \times w \times c}$. To this end, we optimize $\psi_{\mathtt{c}}$ with the cosine distance loss:
\begin{equation}
\label{eq:equ4}
\mathcal{L}_{\mathtt{Log}}= 1 - \cos(X_{\mathtt{L}},X).
\end{equation}
The logical anomaly map $\mathcal{S}_{\mathtt{L}} \in \mathbb{R}^{h \times w}$ is defined as the patch-wise cosine distance between the reconstructed features and the original pre-trained features. Our key insight lies in reconstructing synthesized anomalous features back to pre-trained features rather than their normal discretized counterparts. This provides a continuous learning target that facilitates model convergence and prevents degeneration into trivial solutions. Furthermore, this reconstruction paradigm enhances LogiCo's robustness against segmentation noise; even in the presence of noisy segmentation maps, the CRN can still generate normal reconstructed features to avoid false positive detections.

\subsection{Cross-attentive Structural Reconstruction}

Component-level reconstruction preserves the components' spatial layout, enhancing the granularity of logical anomaly detection. However, due to the lack of intra-component structural information, its capability for structural anomaly detection remains limited. To this end, we introduce a Structural Reconstruction Network (SRN) to reconstruct the fine-grained pre-trained features for structural anomaly detection: $X_{\mathtt{s}} = \psi_{\mathtt{s}}(X) \in \mathbb{R}^{h \times w \times c}$. Similar to the CRN, the loss function is defined as $\mathcal{L}_{\mathtt{Str}}= 1 - \cos(X_{\mathtt{s}},X)$, and the structural anomaly map $\mathcal{S}_{\mathtt{s}} \in \mathbb{R}^{h \times w}$ is computed as the patch-wise cosine distance. The key challenge in the above process is controlling the reconstruction capability of the SRN to prevent it from preserving anomaly cues, which would result in an ``identical shortcut'' \cite{you2022unified, zhang2023exploring, guo2025dinomaly}. To address this, we propose Cross-attentive Structural Reconstruction, a novel method bridging logical and structural anomaly detection. It utilizes the CRN to guide the structural reconstruction process, compelling the SRN to restore anomalous features to their normal representations. Recall that local structural information is discarded during the feature discretization; consequently, the feature space of CRN is free of structural anomalies. Meanwhile, it retains the overall spatial layout of the image, providing a valuable reference for structural reconstruction. Specifically, for the $h$-th intermediate-layer feature $X_{\mathtt{s}}^h \in \mathbb{R}^{hw \times c}$ from the SRN, we extract the corresponding layer feature $X_{\mathtt{L}}^h \in \mathbb{R}^{hw \times c}$ from the CRN to refine the reconstruction process via a cross-attention mechanism:
\begin{align}
\label{eq:equ5}
Q_h=X_{\mathtt{s}}^h W_{Q}^h, \;\; &K_h=X_{\mathtt{L}}^h W_{K}^h, \;\; V_h=X_{\mathtt{L}}^h W_{V}^h, \notag \\
\hat{X}_{\mathtt{s}}^{h}=FFN&(Attention(Q_h,K_h,V_h)),
\end{align}
where $Q_h, K_h, V_h \in \mathbb{R}^{hw \times c}$ denote the query, key, and value matrices, respectively, and $W_{Q}^h, W_{K}^h, W_{V}^h \in \mathbb{R}^{c \times c}$ are the learnable projection weights. $Attention(\cdot)$ and $FFN(\cdot)$ represent the attention operation and the feed-forward network, while $\hat{X}_{\mathtt{s}}^{h}$ is the reconstructed feature map. The above cross-attention operation uses the anomaly-free $X_{\mathtt{L}}^h$ as the key-value pair, explicitly mapping $X_{\mathtt{s}}^h$ to its normal representation, thereby enhancing the reliability of reconstruction while avoiding missed detections. In our implementation, the CRN and SRN adopt Autoencoder and U-Net architectures of the same depth, respectively. Specifically, we insert a self-attention layer into each CRN block to capture global semantics and generate guiding features, and place a cross-attention layer at the corresponding location in the SRN to enable cross-attentive reconstruction. We omit the residual connection in the standard cross-attention module to prevent the leakage of anomalous information.

\subsection{Segmentation Map Discriminator}

The Segmentation Map Discriminator (SMD) is specifically designed to detect quantity-related logical anomalies. We find that learning constraints related to component quantity is challenging in the feature space, whereas segmentation maps offer a more intuitive representation. The SMD directly predicts the anomaly mask based on the segmentation map: $\mathcal{S}_{\mathtt{D}} = \psi_{\mathtt{D}}(G) \in [0,1]^{h \times w}$, where 0 and 1 denote normal and anomalous pixels, respectively. We employ a lightweight fully convolutional U-Net model as our SMD, trained using synthetic segmentation map-mask pairs $(G', \mathcal{M})$, as illustrated in \cref{fig:fig3}. We optimize the SMD using Binary Focal Loss \cite{lin2017focal}, denoted as: 
\begin{equation}
\label{eq:equ6}
\mathcal{L}_{\mathtt{Dis}}= \mathrm{Focal}(\psi_{\mathtt{D}}(G'),\mathcal{M}).
\end{equation}

\subsection{Training and Inference}
We jointly train the different branches within LogiCo, with the overall loss function defined as:
\begin{equation}
\label{eq:equ7}
\mathcal{L} = \mathcal{L}_{\mathtt{Log}} + \mathcal{L}_{\mathtt{Str}} + \mathcal{L}_{\mathtt{Dis}}.
\end{equation}
In the inference phase, we omit the anomaly synthesis and perform the same segmentation and reconstruction steps to obtain anomaly maps $\mathcal{S}_{\mathtt{L}}$, $\mathcal{S}_{\mathtt{s}}$, and $\mathcal{S}_{\mathtt{D}}$ for the query image. Subsequently, we compute the means and standard deviations of these anomaly maps on the validation set, utilizing these statistics for normalization and anomaly score fusion:
\begin{equation}
\label{eq:equ8}
\mathcal{S} = \max\{\frac{\mathcal{S}_{\mathtt{L}}-\mu(\mathcal{S}_{\mathtt{L}})}{\sigma(\mathcal{S}_{\mathtt{L}})}, \frac{\mathcal{S}_{\mathtt{s}}-\mu(\mathcal{S}_{\mathtt{s}})}{\sigma(\mathcal{S}_{\mathtt{s}})}, \frac{\mathcal{S}_{\mathtt{D}}-\mu(\mathcal{S}_{\mathtt{D}})}{\sigma(\mathcal{S}_{\mathtt{D}})}\}.
\end{equation}
To comprehensively capture diverse types of anomalies, we derive the overall anomaly map $\mathcal{S} \in \mathbb{R}^{h \times w}$ by taking the element-wise maximum across these maps. Finally, we upsample $\mathcal{S}$ to the image resolution to obtain the pixel-level anomaly score, while its maximum is used as the image-level anomaly score.

\section{Experiments}

\subsection{Experimental Setup}

\textbf{Datasets.} We conduct extensive evaluations on four anomaly detection benchmarks: MVTec-LOCO \cite{bergmann2022beyond}, MVTec-AD \cite{bergmann2019mvtec}, VisA \cite{zou2022spot}, and Real-IAD \cite{wang2024real}. MVTec-LOCO \cite{bergmann2022beyond} is a benchmark specifically designed for logical anomaly detection, comprising 3,644 images across five categories, each featuring diverse logical and structural anomalies. The MVTec-AD benchmark \cite{bergmann2019mvtec} comprises 5,354 images across 15 industrial products, where certain categories such as ``\texttt{cable}'' and ``\texttt{transistor}'' contain logical anomalies like missing or misplaced components. The VisA benchmark \cite{zou2022spot} comprises 10,821 images across 12 products. Certain categories present significant challenges due to their intricate structures and the presence of multiple target objects. The Real-IAD benchmark \cite{wang2024real} comprises 151,050 multi-view images across 30 different products, with some categories exhibiting subtle structural anomalies that are difficult to detect. We adopt the single-view experimental setting, following recent works \cite{lee2024continuous, liao2025robust, zhang2025towards}.

\textbf{Implementation Details.} We utilize DINOv3 ViT-B/16 \cite{simeoni2025dinov3} as our visual backbone and extract feature maps from the final (12th) layer. All images are resized to a resolution of 512×512 for both semantic segmentation and anomaly detection. For the texture classes in the MVTec-AD dataset \cite{bergmann2019mvtec}, we treat them as a single component and disable the anomaly synthesis and SMD. Otherwise, we uniformly sample from the four anomaly synthesis strategies and apply the same configuration across all other image classes. For the MVTec-AD \cite{bergmann2019mvtec}, VisA \cite{zou2022spot}, and Real-IAD \cite{wang2024real} datasets, we hold out 20\% of the training images to compute the means and standard deviations of the anomaly maps for normalization.

\begin{table}[t]
  \centering
  \renewcommand\arraystretch{1.2}
  \normalsize
  \caption{Comparison of LogiCo with other anomaly detection methods. $\mathcal{y}$ denotes methods specifically designed for logical anomaly detection.}
   \resizebox{\linewidth}{!}{
    \begin{tabular}{ccccccccccccccc}
    \toprule
    \multirow{2}[2]{*}{Method} & \multicolumn{2}{c}{\:MVTec-LOCO\:} & \multicolumn{4}{c}{MVTec-AD} & \multicolumn{4}{c}{VisA}  & \multicolumn{4}{c}{Real-IAD} \\
\cmidrule(lr){2-3}  \cmidrule(lr){4-7}  \cmidrule(lr){8-11} \cmidrule(lr){12-13} \cmidrule{13-15} \multicolumn{1}{c}{} & \multicolumn{1}{c}{{\fontsize{9}{10}\selectfont \:\:I-AUC}\,} & \multicolumn{1}{c}{\,{\fontsize{9}{10}\selectfont sPRO}\:} & \multicolumn{1}{c}{{\fontsize{9}{10}\selectfont \:\:I-AUC}\,} & \multicolumn{1}{c}{\,{\fontsize{9}{10}\selectfont P-AUC}\,} & \multicolumn{1}{c}{\:{\fontsize{9}{10}\selectfont P-AP}\,} & \multicolumn{1}{c}{\:{\fontsize{9}{10}\selectfont PRO\:}\,} & \multicolumn{1}{c}{{\fontsize{9}{10}\selectfont \:\:I-AUC}\,} & \multicolumn{1}{c}{\,{\fontsize{9}{10}\selectfont P-AUC}\,} & \multicolumn{1}{c}{\:{\fontsize{9}{10}\selectfont P-AP}\,} & \multicolumn{1}{c}{\:{\fontsize{9}{10}\selectfont PRO\:}\,} & \multicolumn{1}{c}{{\fontsize{9}{10}\selectfont \:\:I-AUC}\,} & \multicolumn{1}{c}{\,{\fontsize{9}{10}\selectfont P-AUC}\,} & \multicolumn{1}{c}{\:{\fontsize{9}{10}\selectfont P-AP}\,} & \multicolumn{1}{c}{\:{\fontsize{9}{10}\selectfont PRO\,}} \\
    \midrule
    { \fontsize{9}{10}\selectfont PatchCore\cite{roth2022towards}} & 75.6  & 34.4  & 99.1  & 98.4  & 65.8  & 94.4  & 94.9  & 98.5  & 48.0    & 91.0    & 91.7  & 98.9  & 39.6  & 92.8 \\
    { \fontsize{9}{10}\selectfont UniNet\cite{wei2025uninet}} & 83.0    & 65.5  & 99.2  & 98.6  & 64.3  & \underline{95.4}  & 97.9  & 98.7  & 45.6  & 93.0    & 91.7  & 99.2  & 40.9  & 95.8 \\
    { \fontsize{9}{10}\selectfont INP-Former\cite{luo2025exploring}} & 79.3  & 63.0    & \underline{99.5}  & 98.4  & 69.6  & \textbf{95.5} & 98.4  & 98.7  & 49.2  & 93.9  & 93.2  & \underline{99.3}  & \underline{48.4}  & \underline{95.6} \\
    { \fontsize{9}{10}\selectfont Dinomaly\cite{guo2025dinomaly}} & 83.4  & 64.8  & \textbf{99.6} & \textbf{98.8} & \textbf{71.2} & \textbf{95.5} & \textbf{98.9} & \textbf{99.3} & \underline{57.7}  & \underline{94.8}  & \underline{93.7}  & \textbf{99.4} & 48.1  & \textbf{95.7} \\
    { \fontsize{9}{10}\selectfont $\mathcal{y}$\,GLCF\cite{yao2023learning}} & 78.1  & 63.7  & 93.9  & 96.4  & 49.0    & 80.5  & 87.1  & 96.1  & 28.5  & 73.2  & 85.4  & 95.1  & 26.1  & 75.3 \\
    { \fontsize{9}{10}\selectfont $\mathcal{y}$\,AnomalyMoE\cite{gu2025anomalymoe}} & 88.2  & 63.6  & \underline{99.5}  & 98.1  & 67.0    & 94.4  & 98.1  & \underline{99.0}    & 54.6  & \textbf{94.9} & 90.5  & 97.8  & 35.8  & 83.9 \\
    { \fontsize{9}{10}\selectfont $\mathcal{y}$\,CSAD\cite{Hsieh_2024_BMVC}} & 92.9  & \underline{70.6}  & 95.8  & 97.7  & 56.3  & 86.6  & 90.6  & 97.0    & 31.1  & 73.7  & 83.4  & 96.3  & 29.8  & 78.9 \\
    { \fontsize{9}{10}\selectfont $\mathcal{y}$\,SALAD\cite{fuvcka2025salad}} & \underline{93.2}  & 63.3  & 97.2  & 96.9  & 52.6  & 81.1  & 97.4  & 97.4  & 32.7  & 75.7  & 88.8  & 96.2  & 30.2  & 77.9 \\
    {\fontsize{9}{10}\selectfont \cellcolor{gray!15}$\mathcal{y}\,$\textbf{LogiCo}}  & \cellcolor{gray!15}\textbf{96.3} & \cellcolor{gray!15}\textbf{72.9} & \cellcolor{gray!15}\textbf{99.6} & \cellcolor{gray!15}\underline{98.7}  & \cellcolor{gray!15}\underline{70.2}  & \cellcolor{gray!15}95.0    & \cellcolor{gray!15}\underline{98.5}  & \cellcolor{gray!15}\textbf{99.3} & \cellcolor{gray!15}\textbf{60.1} & \cellcolor{gray!15}93.7  & \cellcolor{gray!15}\textbf{93.8} & \cellcolor{gray!15}\textbf{99.4} & \cellcolor{gray!15}\textbf{53.3} & \cellcolor{gray!15}95.3 \\
    \bottomrule
    \end{tabular}%
}
  \label{tab:table1}%
\end{table}%

\begin{table}[t]
  \centering
    \renewcommand\arraystretch{1.0}
    \normalsize
  \caption{Comparison between LogiCo and other logical anomaly detection methods on the MVTec-LOCO dataset. $\mathcal{y}$ indicates that the results are taken from the original paper. LogiCo+ denotes LogiCo using SAM-refined \cite{kirillov2023segment} component segmentation maps.}
     \resizebox{0.7\linewidth}{!}{
    \begin{tabular}{cccccccc}
    \toprule
    \multirow{2}[1]{*}{\;\;{\fontsize{9}{10}\selectfont Method}\;\;} & \multirow{2}[1]{*}{\makecell{\fontsize{9}{10}\selectfont Fine-grained \\ \;\;Segmentation\;\;}} & \multicolumn{3}{c}{\;{\fontsize{9}{10}\selectfont Detection\,(I-AUC)}\;} & \multicolumn{3}{c}{{\fontsize{9}{10}\selectfont Localization\,(sPRO)}} \\
\cmidrule(lr){3-5}  \cmidrule(lr){6-7} \cmidrule{7-8} \multicolumn{1}{c}{} & \multicolumn{1}{c}{} & \multicolumn{1}{c}{\;{\fontsize{9}{10}\selectfont Logical}\;} & \multicolumn{1}{c}{\;{\fontsize{9}{10}\selectfont Structural}\;} & \multicolumn{1}{c}{\;\;{\fontsize{9}{10}\selectfont \textbf{AVG}}\;\;} & \multicolumn{1}{c}{\;{\fontsize{9}{10}\selectfont Logical}\;} & \multicolumn{1}{c}{\;{\fontsize{9}{10}\selectfont Structural}\;} & \multicolumn{1}{c}{\;\;{\fontsize{9}{10}\selectfont \textbf{AVG}}\;\;} \\
    \midrule
    
    {\fontsize{9}{10}\selectfont GLCF\cite{yao2023learning}}& \xmark & 72.9 & 83.2 & 78.1 & 58.7 & 68.6 & 63.7 \\
    {\fontsize{9}{10}\selectfont $\mathcal{y}$GCAD\cite{bergmann2022beyond}}& \xmark & 86.0 &  80.6 & 83.3 & \textbf{71.1} & \underline{69.2} & \underline{70.1} \\
    {\fontsize{9}{10}\selectfont $\mathcal{y}$EfficientAD\cite{batzner2024efficientad}}& \xmark & 86.8 & \underline{94.7} & 90.7 & \multicolumn{1}{c}{-} & \multicolumn{1}{c}{-} & \multicolumn{1}{c}{-} \\
    {\fontsize{9}{10}\selectfont SALAD\cite{fuvcka2025salad}}& \xmark  & \underline{91.7} & \underline{94.7} & \underline{93.2} & 57.4 & \underline{69.2} & 63.3 \\
    {\fontsize{9}{10}\selectfont \cellcolor{gray!15}\textbf{LogiCo}}& \cellcolor{gray!15}\xmark & \cellcolor{gray!15}\textbf{96.5} & \cellcolor{gray!15}\textbf{96.0} & \cellcolor{gray!15}\textbf{96.3} & \cellcolor{gray!15}\underline{63.6} & \cellcolor{gray!15}\textbf{82.2} & \cellcolor{gray!15}\textbf{72.9} \\
    \midrule
    {\fontsize{9}{10}\selectfont PSAD\cite{kim2024few}}&  \cmark & \textbf{97.7} & 91.3  & \underline{94.5} & \multicolumn{1}{c}{-} & \multicolumn{1}{c}{-} & \multicolumn{1}{c}{-} \\
    {\fontsize{9}{10}\selectfont LogSAD\cite{zhang2025logsvd}}& \cmark & 89.3 & 91.6 & 90.4 & 58.3 & 78.8 & 68.6 \\
    {\fontsize{9}{10}\selectfont CSAD\cite{Hsieh_2024_BMVC}}&  \cmark & 92.6 & \underline{93.1} & 92.9 & 65.5 & 75.6 & 70.6 \\
    {\fontsize{9}{10}\selectfont $\mathcal{y}$SAM-LAD\cite{peng2025sam}}&  \cmark & 95.8 & 85.5 & 90.7 & \textbf{85.6} & \underline{80.7} & \textbf{83.2} \\
    {\fontsize{9}{10}\selectfont \cellcolor{gray!15}\textbf{LogiCo+}}& \cellcolor{gray!15}\cmark & \cellcolor{gray!15}\underline{97.1}	& \cellcolor{gray!15}\textbf{96.2} & \cellcolor{gray!15}\textbf{96.7} & \cellcolor{gray!15}\underline{76.2} & \cellcolor{gray!15}\textbf{84.6} & \cellcolor{gray!15}\underline{80.4}\\
    \bottomrule
    \end{tabular}%
    }
  \label{tab:table2}%
\end{table}%

\begin{figure}[t]
  \centering
   \includegraphics[width=0.95\linewidth]{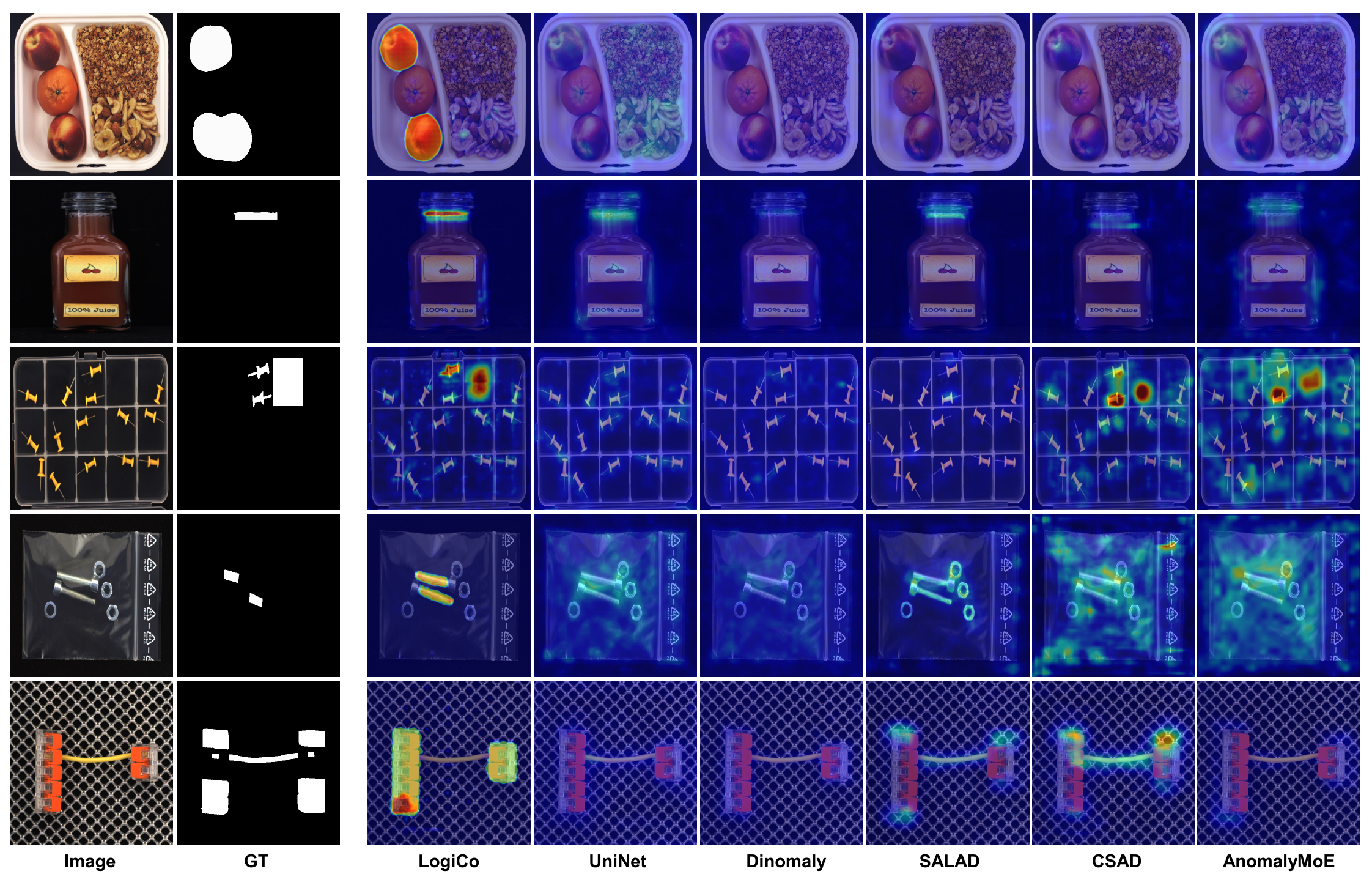}
   \caption{Qualitative comparison of LogiCo and other methods on logical anomaly detection.}
   \label{fig:fig4}
\end{figure}

\textbf{Metrics.} For the MVTec-AD\cite{bergmann2019mvtec}, VisA\cite{zou2022spot}, and Real-IAD\cite{wang2024real} datasets, we evaluate anomaly detection performance using image-level AUROC (I-AUC) and anomaly localization performance using pixel-level AUROC (P-AUC), average precision (P-AP), and per-region overlap (PRO)\cite{bergmann2020uninformed}. For the MVTec-LOCO dataset\cite{bergmann2022beyond}, we employ I-AUC and saturated PRO (sPRO)\cite{bergmann2022beyond} to evaluate anomaly detection and localization performance, respectively, noting that sPRO is specifically designed for logical anomalies. Following previous work\cite{bergmann2022beyond, Hsieh_2024_BMVC, fuvcka2025salad}, metrics for logical and structural anomalies are computed separately, and their average is reported as the overall metric.

\textbf{Baselines.} For structural anomaly detection, we use PatchCore \cite{roth2022towards}, UniNet \cite{wei2025uninet}, INP-Former \cite{luo2025exploring}, and Dinomaly \cite{guo2025dinomaly} as our baseline methods. For logical anomaly detection, we use GLCF \cite{yao2023learning}, AnomalyMoE \cite{gu2025anomalymoe}, CSAD \cite{Hsieh_2024_BMVC}, SALAD \cite{fuvcka2025salad}, GCAD\cite {bergmann2022beyond}, EfficientAD \cite{batzner2024efficientad}, PSAD \cite{kim2024few}, LogSAD \cite{zhang2025logsvd}, and SAM-LAD \cite{peng2025sam} as baseline methods. For INP-Former \cite{luo2025exploring}, Dinomaly \cite{guo2025dinomaly}, and AnomalyMoE \cite{gu2025anomalymoe}, we use the same backbone and image size as LogiCo to ensure a fair comparison. To enable a comprehensive and unified evaluation with consistent metrics and more challenging datasets, we rigorously reproduce the baselines using their official code, with detailed reproduction protocols provided in the \textbf{Appendix}. Unless otherwise specified, all reported results are reproduced results.

\subsection{Experimental Results}

\begin{figure}[t]
  \centering
   \includegraphics[width=0.95\linewidth]{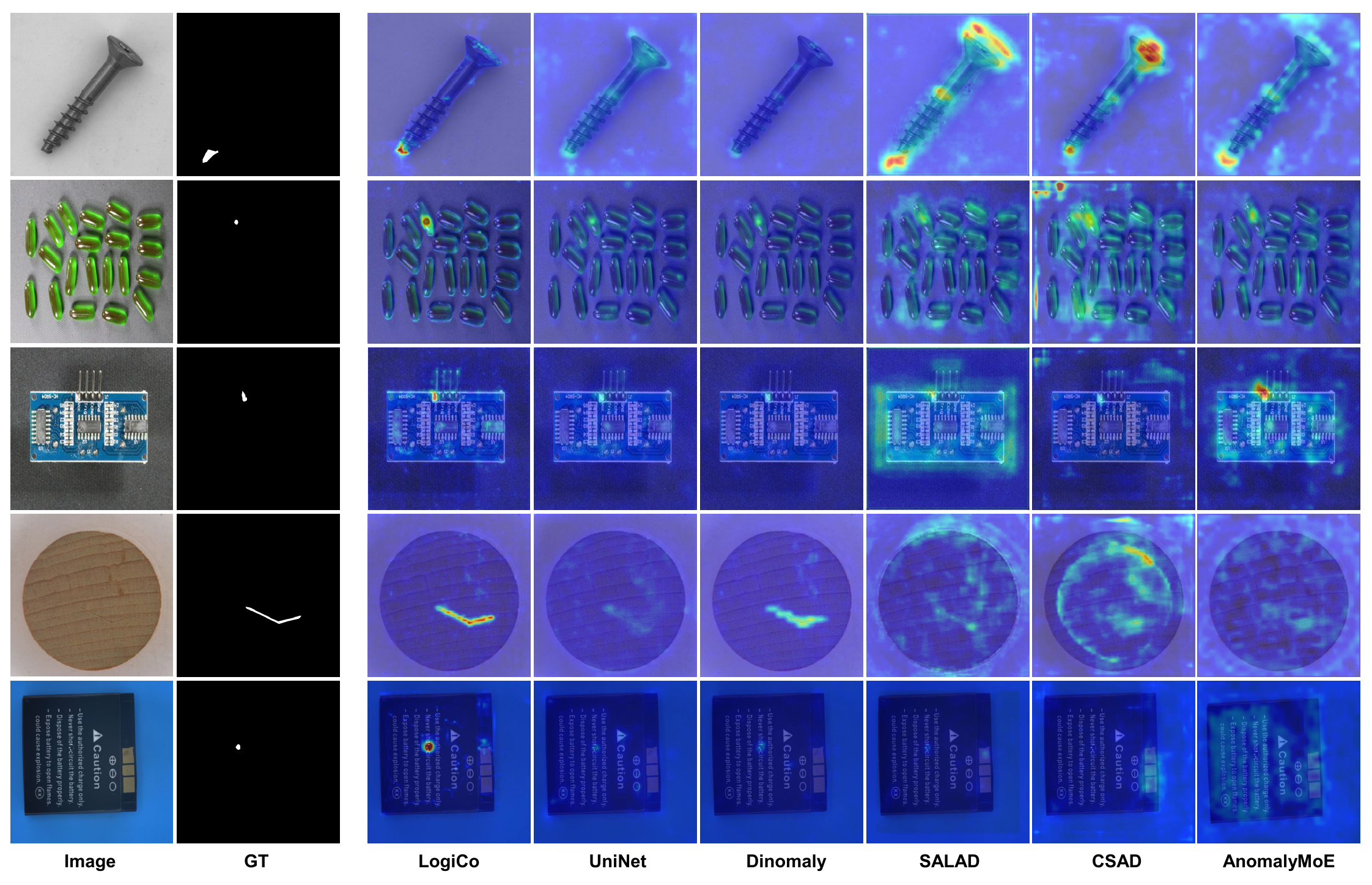}
   \caption{Qualitative comparison of LogiCo and other methods on structural anomaly detection.}
   \label{fig:fig5}
\end{figure}

\begin{figure}[t]
  \centering
   \includegraphics[width=0.97\linewidth]{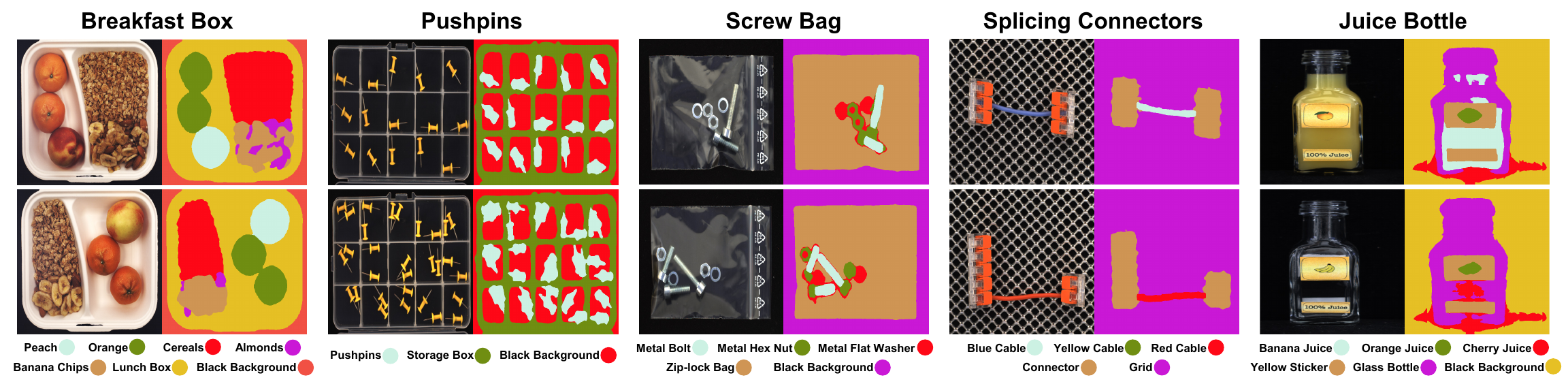}
   \caption{Examples of component segmentation on the MVTec-LOCO dataset.}
   \label{fig:fig6}
\end{figure}

\cref{tab:table1} presents the performance of LogiCo compared to other methods across different benchmarks. Current SOTA methods like Dinomaly\cite{guo2025dinomaly} and INP-Former\cite{luo2025exploring} excel on benchmarks focusing on structural anomalies, such as MVTec-AD\cite{bergmann2019mvtec}, VisA\cite{zou2022spot}, and Real-IAD\cite{wang2024real}. However, they struggle to generalize to the logical anomalies present in the MVTec-LOCO dataset\cite{bergmann2022beyond}. Conversely, methods explicitly designed for logical anomalies, such as CSAD\cite{Hsieh_2024_BMVC}, SALAD\cite{fuvcka2025salad}, and AnomalyMoE\cite{gu2025anomalymoe}, detect logical defects more effectively but lag significantly behind on benchmarks characterized by subtle structural anomalies, particularly on the challenging VisA\cite{zou2022spot} and Real-IAD\cite{wang2024real} datasets. In contrast, LogiCo demonstrates superior performance across all benchmarks. Specifically, it achieves an impressive 96.3\% I-AUC and 72.9\% sPRO on the MVTec-LOCO benchmark\cite{bergmann2022beyond}, while performing on par with or even surpassing top-performing methods on other structural anomaly detection benchmarks. \cref{tab:table2} further reports the logical and structural anomaly detection and localization performance of different methods on the MVTec-LOCO dataset \cite{bergmann2022beyond}. Among them, some approaches \cite{kim2024few, fuvcka2025salad} achieve strong anomaly detection performance due to their global semantic modeling capabilities, but perform poorly in anomaly localization. In contrast, some methods \cite{Hsieh_2024_BMVC, peng2025sam} excel at anomaly localization by leveraging fine-grained segmentation maps, yet show limited performance in anomaly detection. By comparison, LogiCo achieves a favorable trade-off between these two aspects. Moreover, when using finer segmentation maps, LogiCo obtains a substantial performance improvement. Figs. \ref{fig:fig4} and \ref{fig:fig5} present the qualitative results of LogiCo and other methods on logical and structural anomalies, respectively. The results demonstrate that LogiCo precisely localizes various types of anomaly regions, significantly extending the detectable range of existing methods and emphasizing its feasibility for real-world scenarios.

\begin{table}[t]
  \centering
  \normalsize
  \renewcommand\arraystretch{1.}
  \caption{Comparison of computational efficiency between LogiCo and other methods.}
     \resizebox{0.7\linewidth}{!}{
    \begin{tabular}{cccc}
    \toprule
    \multicolumn{1}{c}{\multirow{1}[4]{*}{{\fontsize{9}{10}\selectfont Method}}} & \multicolumn{1}{c}{\multirow{1}[4]{*}{\:\:{\fontsize{9}{10}\selectfont \makecell{Training \\ Time (h) $\downarrow$}}\:\:}} & \multicolumn{1}{c}{\multirow{1}[4]{*}{\:\:{\fontsize{9}{10}\selectfont Images/s $\uparrow$ }\:\:}} & {\fontsize{9}{10}\selectfont Performance on MVTec-LOCO $\uparrow$} \\
    \multicolumn{1}{c}{} &       &       & {\fontsize{7}{10}\selectfont \:(Log.-AUC/Str.-AUC/Log.-sPRO/Str.-sPRO)\:} \\
    \midrule
    {\fontsize{8}{10}\selectfont Dinomaly\cite{guo2025dinomaly}} & 0.7   & 30.2       & 73.71 / 93.04 / 45.83 / \textbf{83.74} \\
    {\fontsize{8}{10}\selectfont AnomalyMoE\cite{gu2025anomalymoe}} & 18.0    & 15.2  & 82.79 / 93.57 / 58.36 / 68.78 \\
    {\fontsize{8}{10}\selectfont SALAD\cite{fuvcka2025salad}} & 1.7   & 37.6 & 91.73 / 94.69 / 57.39 / 69.22 \\
    {\fontsize{8}{10}\selectfont LogSAD\cite{zhang2025logsvd}} & 0.2 & 0.3   & 89.29 / 91.58 / 58.31 / 78.81 \\
    {\fontsize{8}{10}\selectfont \cellcolor{gray!15}\textbf{LogiCo}} & \cellcolor{gray!15}1.0     & \cellcolor{gray!15}17.4  & \cellcolor{gray!15}\textbf{96.53}\,/\,\textbf{96.03}\,/\,\textbf{63.63}\,/ 82.16 \\
    \bottomrule
    \end{tabular}%
    }
  \label{tab:table3}%
\end{table}%

\begin{table}[t]
  \centering
  \renewcommand\arraystretch{1.0}
  \caption{Ablation study results of LogiCo components on the MVTec-LOCO dataset.}
  \resizebox{0.7\linewidth}{!}{
    \begin{tabular}{cccccccc}
    \toprule
    \multicolumn{1}{c}{\multirow{2}[1]{*}{{\fontsize{8}{10}\selectfont \:CRN\:}}} & \multicolumn{1}{c}{\multirow{2}[1]{*}{{\fontsize{8}{10}\selectfont \:SRN\:}}} & \multicolumn{1}{c}{\multirow{2}[1]{*}{{\fontsize{8}{10}\selectfont \:SMD\:}}} & \multicolumn{1}{c}{\multirow{2}[1]{*}{{\fontsize{8}{10}\selectfont \:Cross-Attn.\:}}} & \multicolumn{2}{c}{{\fontsize{8}{10}\selectfont \:\:Detection\,(I-AUC)\:\:}} & \multicolumn{2}{c}{{\fontsize{8}{10}\selectfont \:\:Localization\,(sPRO)\:}} \\
        \cmidrule(lr){5-6}    \cmidrule(lr){7-8}  \cmidrule{8-8}     &       &       &       & \multicolumn{1}{c}{{\fontsize{8}{10}\selectfont \:\:Logical\:}} & \multicolumn{1}{c}{{\fontsize{8}{10}\selectfont \:Structural\:}} & \multicolumn{1}{c}{{\fontsize{8}{10}\selectfont \:\:\:Logical\:}} & \multicolumn{1}{c}{{\fontsize{8}{10}\selectfont \:Structural\:}} \\
    \midrule
         \cmark & \multicolumn{1}{c}{-} & \multicolumn{1}{c}{-} & \multicolumn{1}{c}{-} & 90.88 & 84.59 & 67.51 & 70.07 \\
    \multicolumn{1}{c}{-} &    \cmark   & \multicolumn{1}{c}{-} & \multicolumn{1}{c}{-} & 75.34 & 94.86 & 53.00    & 81.55 \\
    \multicolumn{1}{c}{-} & \multicolumn{1}{c}{-} &    \cmark   & \multicolumn{1}{c}{-} & 85.63 & 61.20  & 58.05 & 35.21 \\
         \cmark & \multicolumn{1}{c}{-} &  \cmark    & \multicolumn{1}{c}{-} & \textbf{96.87} & 82.47 & 64.22 & 62.84 \\
    \multicolumn{1}{c}{-} &   \cmark    &   \cmark    & \multicolumn{1}{c}{-} & 86.85 & 93.84 & 56.79 & 80.00 \\
        \cmark   &    \cmark   & \multicolumn{1}{c}{-} & \multicolumn{1}{c}{-} & 89.43 & 94.94 & \underline{67.71} & 81.98 \\
        \cmark  &   \cmark    & \multicolumn{1}{c}{-} &   \cmark    & 90.05 & \underline{95.95} & \textbf{68.08} & \textbf{84.71} \\
        \cmark  &   \cmark    &     \cmark  & \multicolumn{1}{c}{-} & 95.81 & 94.82 & 63.04 & 81.05 \\
        \cellcolor{gray!15}\cmark  &   \cellcolor{gray!15}\cmark    &   \cellcolor{gray!15}\cmark    &    \cellcolor{gray!15}\cmark   & \cellcolor{gray!15}\underline{96.53} & \cellcolor{gray!15}\textbf{96.03} & \cellcolor{gray!15}63.63 & \cellcolor{gray!15}\underline{82.16} \\
    \bottomrule
    \end{tabular}%
    }
  \label{tab:table4}%
\end{table}%

\cref{fig:fig6} displays the component segmentation results of LogiCo on the MVTec-LOCO dataset. Unlike prior works that rely on elaborate segmentation pipelines \cite{Hsieh_2024_BMVC, kim2024few, zhang2025logsvd, peng2025sam}, LogiCo adopts a simple open-vocabulary segmentation scheme, significantly reducing computational overhead. Moreover, LogiCo demonstrates robustness against segmentation noise. Specifically, in the \texttt{Juice Bottle} category, it yields an I-AUC of 98.93\% and an sPRO of 82.67\% (refer to the \textbf{Appendix} for per-category metrics), despite the segmentation module struggling to distinguish between specific juice types. We attribute this to LogiCo's robust reconstruction capability, which effectively restores normal feature representations from noisy inputs, thereby mitigating false positives. More experimental results and analysis on component segmentation are provided in the \textbf{Appendix}.

\cref{tab:table3} compares the computational efficiency of different methods. The experiments are implemented in PyTorch and conducted on a single Nvidia RTX 5090 GPU. At inference time, we set the batch size to 1 to mimic practical deployment scenarios. Benefiting from a simplified component segmentation pipeline, LogiCo eliminates the need to train an additional component segmentation model \cite{fuvcka2025salad, Hsieh_2024_BMVC, kim2024few}, which shortens its training time and enables it to meet the practical demands of rapid product iteration and updates. Furthermore, LogiCo achieves a high inference speed of 17.4 FPS at a resolution of $512 \times 512$, thereby effectively balancing anomaly detection performance and efficiency to meet the requirements of real-world scenarios.

\begin{table}[t]
  \centering
  \renewcommand\arraystretch{1.0}
  \caption{Ablation study results of anomaly synthesis on the MVTec-LOCO dataset.}
    \resizebox{0.7\linewidth}{!}{
    \begin{tabular}{ccccc}
    \toprule
    \multirow{2}[2]{*}{\makecell{{\fontsize{9}{10}\selectfont Anomaly Synthesis} \\ {\fontsize{9}{10}\selectfont Method}}} & \multicolumn{2}{c}{{\fontsize{8}{10}\selectfont Detection\;(I-AUC)}} & \multicolumn{2}{c}{{\fontsize{8}{10}\selectfont \:\:Localization\;(sPRO)}} \\
\cmidrule(lr){2-3}  \cmidrule(lr){4-5} \cmidrule{5-5} \multicolumn{1}{c}{} & \multicolumn{1}{c}{{\fontsize{8}{10}\selectfont \:Logical\:}} & \multicolumn{1}{c}{{\fontsize{8}{10}\selectfont \:Structural\:}} & \multicolumn{1}{c}{{\fontsize{8}{10}\selectfont \:\:Logical\:}} & \multicolumn{1}{c}{{\fontsize{8}{10}\selectfont \:\:Structural\:}} \\
    \midrule
    {\fontsize{8}{10}\selectfont \textbf{Deletion}} & 85.70  & 94.64 & 46.03 & 81.13 \\
    {\fontsize{8}{10}\selectfont \textbf{Replacement}} & 88.47 & 95.44 & 53.08 & 82.04 \\
    {\fontsize{8}{10}\selectfont \textbf{Erasure}} & 90.29 & 95.81 & 52.81 & \underline{82.37} \\
    {\fontsize{8}{10}\selectfont \textbf{Addition}}   & 93.41 & 95.01 & 61.45 & 81.99 \\
    {\fontsize{8}{10}\selectfont \textbf{Addition}$+$\textbf{Deletion}} & 93.73 & 95.24 & 61.90  & 82.12 \\
    {\fontsize{8}{10}\selectfont \textbf{Addition}$+$\textbf{Replacement}} & 94.58 & 95.35 & 62.68 & 82.09 \\
    {\fontsize{8}{10}\selectfont \textbf{Addition}$+$\textbf{Erasure}} & 95.73 & 95.59 & 62.29 & \textbf{82.39} \\
    {\fontsize{8}{10}\selectfont \:\textbf{Addition}$+$\textbf{Replacement}$+$\textbf{Deletion}\:} & 94.88 & 95.55 & 62.84 & 81.89 \\
    {\fontsize{8}{10}\selectfont \textbf{Addition}$+$\textbf{Erasure}$+$\textbf{Deletion}} & 95.99 & 95.72 & 63.36 & 82.08 \\
    {\fontsize{8}{10}\selectfont \textbf{Addition}$+$\textbf{Erasure}$+$\textbf{Replacement}} & \underline{96.26} & \underline{95.91} & \textbf{63.65} & 82.13 \\
    {\fontsize{8}{10}\selectfont \cellcolor{gray!15}\textbf{All}}   & \cellcolor{gray!15}\textbf{96.53} & \cellcolor{gray!15}\textbf{96.03} & \cellcolor{gray!15}\underline{63.63} & \cellcolor{gray!15}82.16 \\
    \bottomrule
    \end{tabular}%
    }
  \label{tab:table5}%
\end{table}%

\begin{table}[t]
  \centering
  \normalsize
  \renewcommand\arraystretch{1.0}
  \caption{Ablation study results on different backbones and image sizes.}
    \resizebox{0.7\linewidth}{!}{
    \begin{tabular}{cccccccc}
    \toprule
    \multirow{2}[1]{*}{{\fontsize{9}{10}\selectfont Backbone}} & \multicolumn{1}{c}{\multirow{2}[1]{*}{{\fontsize{9}{10}\selectfont \:\:Image Size\:\:}}} & \multicolumn{2}{c}{{\fontsize{9}{10}\selectfont \:MVTec-LOCO\:}} & \multicolumn{4}{c}{\fontsize{9}{10}\selectfont MVTec-AD} \\
\cmidrule(lr){3-4} \cmidrule(lr){5-7} \cmidrule{6-8} \multicolumn{1}{c}{} &       & \multicolumn{1}{c}{{\fontsize{9}{10}\selectfont \:\:I-AUC\:\:}} & \multicolumn{1}{c}{{\fontsize{9}{10}\selectfont \:\:sPRO\:\:}} & \multicolumn{1}{c}{{\fontsize{9}{10}\selectfont\:\:I-AUC\:\:}} & \multicolumn{1}{c}{{\fontsize{9}{10}\selectfont \:\:P-AUC\:\:}} & \multicolumn{1}{c}{{\fontsize{9}{10}\selectfont \:\:P-AP\:\:}} & \multicolumn{1}{c}{{\fontsize{9}{10}\selectfont\:\:PRO\:\:}} \\
    \midrule
    \multirow{2}[1]{*}{\fontsize{9}{10}\selectfont DINOv2-R\cite{darcet2024vision}} & 224$^2$   & 94.42 & 69.48 & 99.30  & 98.14 & 65.98 & 93.19 \\
    \multicolumn{1}{c}{} & 448$^2$   & 95.67 & 71.85 & 99.55 & 98.58 & 68.09 & 94.27 \\
    \midrule
    {\fontsize{9}{10}\selectfont DINOv3{$_{\text{ViT-S}}$}\cite{simeoni2025dinov3}} & 512$^2$   & 95.50  & 70.20  & 99.49 & 98.43 & 68.81 & 94.37 \\
    \midrule
    \multirow{1}[7]{*}{\fontsize{9}{10}\selectfont DINOv3{$_{\text{ViT-B}}$}\cite{simeoni2025dinov3}} & 256$^2$   & 94.78 & 69.65 & 99.04 & 98.33 & 66.81 & 93.75 \\
    \multicolumn{1}{c}{} & 384$^2$   & 95.59 & 71.38 & 99.40  & 98.47 & 68.30  & 94.26 \\
    \multicolumn{1}{c}{} & \cellcolor{gray!15}512$^2$   & \cellcolor{gray!15}\textbf{96.28} & \cellcolor{gray!15}\underline{72.90}  & \cellcolor{gray!15}\underline{99.59} & \cellcolor{gray!15}\underline{98.68} & \cellcolor{gray!15}\underline{70.19} & \cellcolor{gray!15}\underline{95.00} \\
    \midrule
    {\:\:\fontsize{9}{10}\selectfont DINOv3{$_{\text{ViT-L}}$}\cite{simeoni2025dinov3}\:\:} & 512$^2$   & \underline{95.80}  & \textbf{73.88} & \textbf{99.70} & \textbf{98.73} & \textbf{70.98} & \textbf{95.29} \\
    \bottomrule
    \end{tabular}%
    }
  \label{tab:table6}%
\end{table}%

\subsection{Ablation Study}

\textbf{Ablation Study on LogiCo Components.} \cref{tab:table4} validates the effectiveness of each component within LogiCo. As a key module of LogiCo, the CRN provides the primary capability for logical anomaly detection. When used independently, it achieves a logical-AUC of 90.88\% and a logical-sPRO of 67.51\%, which outperforms a wide range of logical anomaly detection methods \cite{bergmann2022beyond, yao2023learning, batzner2024efficientad, zhang2025logsvd}. In contrast, the SRN is mainly responsible for detecting structural anomalies, compensating for CRN's limitations in this aspect. The introduction of the SRN improves the structural anomaly detection scores by approximately 10\%. Furthermore, by employing cross-attentive reconstruction, SRN's anomaly detection and localization performance is further enhanced, with structural AUC and sPRO increasing by 1.01\% and 2.73\%, respectively, thereby demonstrating its indispensability. The SMD module is primarily designed to identify quantity-related logical anomalies, thereby expanding the detection scope of LogiCo. However, incorporating SMD may compromise anomaly localization performance. Since it operates on segmentation maps devoid of appearance details, it can effectively identify the anomalous component but fails to pinpoint the precise location of the anomaly within it. Despite this limitation, we advocate using SMD for products subject to quantity constraints to improve overall detection performance.

\textbf{Ablation Study on Anomaly Synthesis.} \cref{tab:table5} investigates the impact of different anomaly synthesis methods. Compared with using a single anomaly synthesis method, using multiple strategies simultaneously leads to significant performance improvements. Furthermore, for the MVTec-LOCO dataset, the \textbf{Addition} and \textbf{Erasure} operations demonstrate superior performance, which may be attributed to their ability to generate diverse anomalies. Our anomaly synthesis process operates within a discrete pixel space, facilitating straightforward customization tailored to the logical constraints of specific products. This flexibility significantly enhances LogiCo's adaptability and scalability across diverse application scenarios.

\textbf{Ablation Study on Feature Extraction.} \cref{tab:table6} reports the anomaly detection performance of LogiCo using different backbones and image sizes. When using the popular DINOv2-R backbone\cite{darcet2024vision}, LogiCo consistently delivers excellent results, confirming its broad effectiveness across various pre-trained models. Furthermore, we observe that performance improves notably as the backbone size and input image resolution increase; however, this incurs a higher computational overhead. In this paper, we employ DINOv3 ViT-B \cite{simeoni2025dinov3} with a resolution of $512 \times 512$ as the default configuration to balance performance and computational efficiency. \cref{tab:table7} investigates the impact of extracting pre-trained features from different intermediate layers, using DINOv3 ViT-B as the backbone. For multi-layer settings, we reconstruct the averaged feature representation. We observe that LogiCo benefits from feature maps at deeper layers; specifically, utilizing late layers improves anomaly detection performance by 2.37\% on MVTec-LOCO and 0.38\% on MVTec-AD compared to early layers. It is worth noting that LogiCo achieves the best overall performance when using only the final layer. This can be ascribed to the Gram Anchoring technique \cite{simeoni2025dinov3} employed by DINOv3, which imposes additional consistency constraints on the final layer during pre-training, enabling it to generate higher-quality feature maps.

\begin{table}[t]
  \centering
  \renewcommand\arraystretch{1.1}
  \caption{Ablation study results on different intermediate layers.}
   \resizebox{0.7\linewidth}{!}{
    \begin{tabular}{ccccccc}
    \toprule
    \multirow{2}[2]{*}{\fontsize{11}{11}\selectfont Layers} & \multicolumn{2}{c}{\fontsize{9}{10}\selectfont \:\:MVTec-LOCO\:\:} & \multicolumn{4}{c}{\fontsize{9}{10}\selectfont \:\:MVTec-AD\:\:} \\
\cmidrule(lr){2-3} \cmidrule(lr){4-6}   \cmidrule{5-7} \multicolumn{1}{c}{} & \multicolumn{1}{c}{\fontsize{9}{10}\selectfont \:\:I-AUC\:\:} & \multicolumn{1}{c}{\fontsize{9}{10}\selectfont \:\:sPRO\:\:} & \multicolumn{1}{c}{\fontsize{9}{10}\selectfont \:I-AUC\:} & \multicolumn{1}{c}{\fontsize{9}{10}\selectfont \:P-AUC\:} & \multicolumn{1}{c}{\fontsize{9}{10}\selectfont \:P-AP\:} & \multicolumn{1}{c}{\fontsize{9}{10}\selectfont \:PRO\:} \\
    \midrule
    {{\fontsize{9}{10}\selectfont \:Early Layers}$_{ \mathtiny{\{1,2,3,4,5,6\}}}$\:} & 93.33 & 67.15 & 99.19 & 98.00   & 65.18 & 94.14 \\
    {{\fontsize{9}{10}\selectfont \:Middle Layers}$_{\mathtiny{\{4,5,6,7,8,9\}}}$\:}  & 93.51 & 68.83 & 99.40  & 98.20  & 68.84 & \textbf{95.38} \\
    {{\fontsize{9}{10}\selectfont \:Late Layers}$_{\mathtiny{\{7,8,9,10,11,12\}}}$\:}   & \underline{95.70}  & \underline{71.24} & \underline{99.57} & 98.36 & 69.03 & \underline{95.26} \\
     {{\fontsize{9}{10}\selectfont Even-numbered Layers}$_{\mathtiny{\{2,4,6,8,10,12\}}}$}   & 95.28 & 70.68 & 99.56 & \underline{98.52} & \textbf{70.34} & \underline{95.26} \\
    \cellcolor{gray!15}{{\fontsize{9}{10}\selectfont \:\:Final Layer}$_{\mathtiny{\{12\}}}$\:\:} & \cellcolor{gray!15}\textbf{96.28} & \cellcolor{gray!15}\textbf{72.90} & \cellcolor{gray!15}\textbf{99.59} & \cellcolor{gray!15}\textbf{98.68} & \cellcolor{gray!15}\underline{70.19} & \cellcolor{gray!15}95.00 \\
    \bottomrule
    \end{tabular}%
    }
  \label{tab:table7}%
\end{table}%

\section{Conclusion}

In this paper, we proposed LogiCo, a unified framework for logical and structural anomaly detection. LogiCo learns diverse logical constraints via component-level feature reconstruction, and extends its detection capability by integrating cross-attentive structural reconstruction with segmentation map discrimination, effectively addressing the limited detection scope and granularity of existing methods. Specifically, LogiCo comprises a component-level reconstruction network, a structural reconstruction network, and a segmentation map discriminator, all operating in a fine-grained discriminative space, enabling effective detection of diverse anomalies. Extensive experiments on multiple benchmarks demonstrate that LogiCo achieves SOTA performance in both logical and structural anomaly detection. In future work, we aim to further enhance LogiCo's adaptability across diverse application scenarios, specifically by developing learnable anomaly synthesis strategies and optimization mechanisms.

\bibliographystyle{splncs04}
\bibliography{main}

\clearpage

\renewcommand*{\thefigure}{S\arabic{figure}}
\renewcommand*{\thetable}{S\arabic{table}}
\renewcommand*{\theequation}{S\arabic{equation}}
\renewcommand\thesection{\Alph{section}}
\setcounter{table}{0}
\setcounter{figure}{0}
\setcounter{equation}{0}
\setcounter{section}{0}

\title{LogiCo: A Unified Framework for Logical and Structural Anomaly Detection \\ \textit{Supplementary Material}} 

% TODO REVIEW: If the paper title is too long for the running head, you can set
% an abbreviated paper title here. If not, comment out.
\titlerunning{LogiCo: A Unified Framework for Logical and Structural Anomaly Detection}

% TODO FINAL: Replace with your author list. 
% Include the authors' OCRID for the camera-ready version, if at all possible.
\author{Ximiao Zhang\inst{1} \and
Min Xu\inst{2} \and
Xiuzhuang Zhou\inst{1}\textsuperscript{\Letter}}

% TODO FINAL: Replace with an abbreviated list of authors.
\authorrunning{X. Zhang et al.}
% First names are abbreviated in the running head.
% If there are more than two authors, 'et al.' is used.

% TODO FINAL: Replace with your institution list.
% TODO FINAL: Replace with your institution list.
\institute{School of Intelligent Engineering and Automation, Beijing University of Posts and Telecommunications \\ \email{xiuzhuang.zhou@bupt.edu.cn}\and
College of Information and Engineering, Capital Normal University}

\maketitle
\appendix
\setcounter{footnote}{0}

\section{Reproduction Protocols}

This section introduces our reproduction protocols. It should be noted that, for all ViT-based methods, we set the input resolution to 512 $\times$ 512 to ensure consistency with LogiCo. For CNN-based methods, we keep their default input resolutions unchanged, since we observe that increasing the resolution for such methods can instead lead to performance degradation, as also reported in prior works\cite{luo2025exploring, guo2025dinomaly}.

\noindent  \textbf{PatchCore}\cite{roth2022towards}: We use the official implementation of Roth et al.\footnote{\url{https://github.com/amazon-science/patchcore-inspection}}, with WRN-50 as the backbone, extracting features from the second and third intermediate layers. All images are resized to a resolution of 256 $\times$ 256, and the subsampling ratio is set to 0.1.

\noindent \textbf{UniNet}\cite{wei2025uninet}: We use the official implementation of Wei et al.\footnote{\url{https://github.com/pangdatangtt/UniNet}}, with WRN-50 as the backbone, and reconstruct the first three intermediate layers. All images are resized to a resolution of 256 $\times$ 256.

\noindent \textbf{GLCF}\cite{yao2023learning}: We use the official implementation of Yao et al.\footnote{\url{https://github.com/hmyao22/GLCF}}, and all experimental settings follow the original paper.

\noindent \textbf{INP-Former}\cite{luo2025exploring}: We use the official implementation of Luo et al.\footnote{\url{https://github.com/luow23/INP-Former}} under the single-class experimental setting. We employ DINOv3-ViT-B/16 as the backbone and set the input image resolution to 512 $\times$ 512, while keeping all other settings consistent with the original paper. 

\noindent \textbf{Dinomaly}\cite{guo2025dinomaly}: We use the official implementation of Guo et al.\footnote{\url{https://github.com/guojiajeremy/Dinomaly}} under the single-class experimental setting. We employ DINOv3-ViT-B/16 as the backbone and set the input image resolution to 512 $\times$ 512, while keeping all other settings consistent with the original paper. 

\noindent \textbf{AnomalyMoE}\cite{gu2025anomalymoe}: We use the official implementation of Gu et al.\footnote{\url{https://github.com/CASIA-LMC-Lab/AnomalyMoE}} under the single-class experimental setting. We employ DINOv3-ViT-B/16 as the backbone for feature extraction and component segmentation clustering, with the number of clusters set to 5.

\noindent \textbf{CSAD}\cite{Hsieh_2024_BMVC}: We use the official implementation of Hsieh et al.\footnote{\url{https://github.com/Tokichan/CSAD}} This method adopts a multi-stage pipeline. Specifically, we first generate pseudo-labels using the component segmentation pipeline from CSAD, and then train the Component Segmentation Network with these pseudo-labels. The trained Component Segmentation Network is then used to predict global anomaly detection results, which are further aggregated with the outputs of the reconstruction branch to obtain the final anomaly map.

\noindent \textbf{SALAD}\cite{fuvcka2025salad}: We use the official implementation of Fu{\v c}ka et al.\footnote{\url{https://github.com/MaticFuc/SALAD}} The overall reproduction procedure is consistent with CSAD, and we use the same parameter settings as reported in the original paper.

\noindent \textbf{PSAD}\cite{kim2024few}: We use the official implementation of Kim et al.\footnote{\url{https://github.com/oopil/PSAD_logical_anomaly_detection}} We use the few-shot annotated segmentation masks provided by the authors to train the component segmentation model. Following the original experimental setup, we employ PatchCore \cite{roth2022towards} with a WRN-101 backbone to predict multi-scale anomaly scores.

\noindent \textbf{LogSAD}\cite{zhang2025logsvd}: We use the official implementation of Zhang et al.\footnote{\url{https://github.com/zhang0jhon/LogSAD}} and adopt the full-data protocol, with all experimental settings following the original paper.

For the other baselines, such as \textbf{GCAD} \cite{bergmann2022beyond}, \textbf{EfficientAD} \cite{batzner2024efficientad}, and \textbf{SAM-LAD} \cite{peng2025sam}, since no official implementations are available, we directly adopt the results reported in their original papers.

\begin{table}[h]
  \centering
  \renewcommand\arraystretch{1.1}
  \caption{Ablation study results on component-level reconstruction targets.}
    \resizebox{0.6\linewidth}{!}{
    \begin{tabular}{ccccc}
    \toprule
    \multicolumn{1}{c}{\multirow{2}[2]{*}{\makecell{\:\:Reconstruction\:\: \\ Target}}} & \multicolumn{2}{c}{Detection (I-AUC)} & \multicolumn{2}{c}{Localization (sPRO)} \\
\cmidrule(lr){2-3}  \cmidrule(lr){4-5} \cmidrule{4-5}         & \multicolumn{1}{c}{\:\:Logical\:\:} & \multicolumn{1}{c}{\:\:Structural\:\:} & \multicolumn{1}{c}{\:\:Logical\:\:} & \multicolumn{1}{c}{\:\:Structural\:\:} \\
    \midrule
     $\psi_{\mathtt{c}}(X'_\mathtt{c})\rightarrow X_\mathtt{c}$     & 82.26 & 66.27 & 56.53 & 24.51 \\
     \cellcolor{gray!15}$\psi_{\mathtt{c}}(X'_\mathtt{c})\rightarrow X$     & \cellcolor{gray!15}\textbf{90.88} & \cellcolor{gray!15}\textbf{84.59} & \cellcolor{gray!15}\textbf{67.51} & \cellcolor{gray!15}\textbf{70.07} \\
    \bottomrule
    \end{tabular}%
    }
  \label{tab:tables1}%
\end{table}%

\section{More Results}

This section presents additional experimental results, including ablation studies on component-level reconstruction targets and count-related anomalies, further experiments and analysis on robustness to segmentation noise, supplementary qualitative results, and per-category metrics.

\textbf{Ablation Study on Component-level Reconstruction Targets.} \cref{tab:tables1} investigates the impact of different reconstruction targets on the performance of the Component-level Reconstruction Network (CRN). We observe a significant performance degradation when the CRN is trained to reconstruct the original discretized representations ($X_\mathtt{c}$) from synthesized anomalous features. This reconstruction objective risks collapsing to a trivial identity mapping, where the network simply retains input features to minimize the loss rather than learning selective reconstruction, thereby hindering generalization. Furthermore, this strategy is highly sensitive to the quality of component segmentation, leading to an increased false-positive rate when the segmentation maps are noisy.

\textbf{Ablation Study on Count-Related Anomalies.} \cref{tab:tables2} validates the effectiveness of the Segmentation Map Discriminator (SMD) in detecting count-related anomalies. We independently evaluate three types of anomalies in the ``Breakfast Box'' category of the MVTec-LOCO dataset \cite{bergmann2022beyond}. The experimental results demonstrate that SMD is effective in identifying count-related anomalies.

\begin{table}[t]
  \centering
  \renewcommand\arraystretch{1.1}
  \caption{Comparison of I-AUC with and without SMD.}
  \resizebox{0.6\linewidth}{!}{
   \scriptsize
    \begin{tabular}{c|ccc}
    \toprule
    \multicolumn{1}{c|}{Setting} & Count & Misplacement & Missing \\
    \midrule
    w/o SMD & \multicolumn{1}{c}{90.08} & \multicolumn{1}{c}{98.63} & \multicolumn{1}{c}{99.33} \\
    w/ SMD & 99.46\textcolor{myred}{$\mathttt{(9.38\uparrow)}$} & 98.96\textcolor{mygreen}{$\mathttt{(0.33\uparrow)}$} & 99.98\textcolor{mygreen}{$\mathttt{(0.65\uparrow)}$} \\
    \bottomrule
    \end{tabular}%
    }
  \label{tab:tables2}%
  \vspace{0.3cm}
\end{table}%

\begin{figure}[t]
  \centering
   \includegraphics[width=\linewidth]{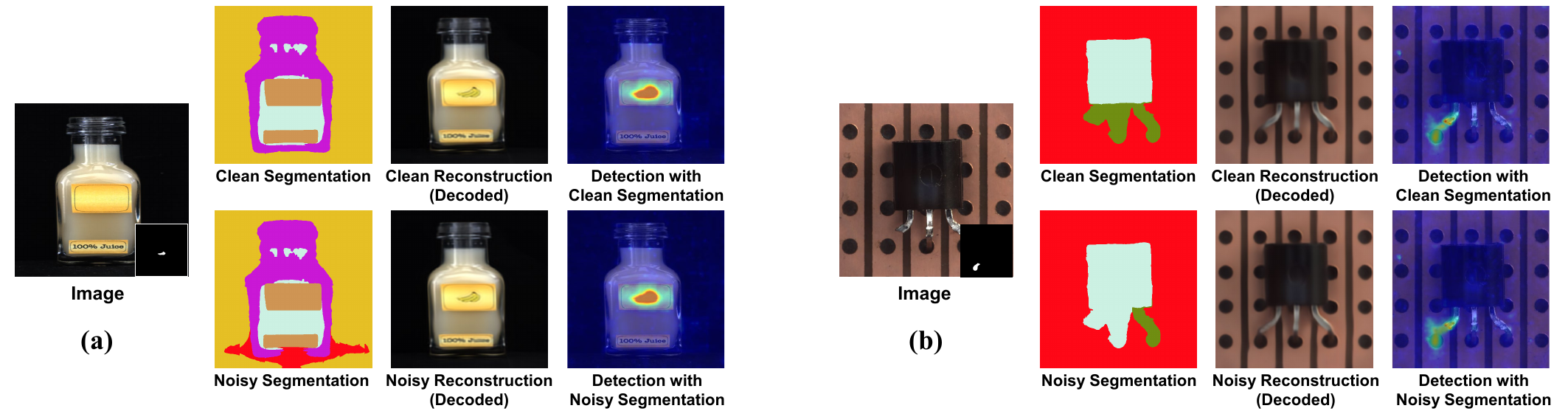}
   \caption{Component-level reconstruction and anomaly detection results of LogiCo under clean and noisy segmentation maps.}
   \label{fig:figs1}
\end{figure}

\textbf{Robustness to Segmentation Noise.} \cref{fig:figs1} shows the component-level reconstruction and anomaly detection results of LogiCo under both clean and noisy segmentation settings. In \cref{fig:figs1}(a), the segmentation module incorrectly classifies the bottom region of the juice bottle as ``\textit{cherry juice}''. However, despite the mismatch between the segmentation map and the input image, our reconstruction network still successfully restores the noisy discretized feature maps to their normal pre-trained representations, thereby avoiding false positive detections. \cref{fig:figs1}(b) presents a more extreme example where the ``\textit{metal pins}'' are misidentified as the body part of the transistor. Nevertheless, the reconstruction network yields reconstruction results identical to those obtained with clean segmentation. This confirms that our approach effectively mitigates the dependency on precise segmentation, validating its robustness even in non-ideal conditions.

\begin{figure}[t]
  \centering
   \includegraphics[width=0.93\linewidth]{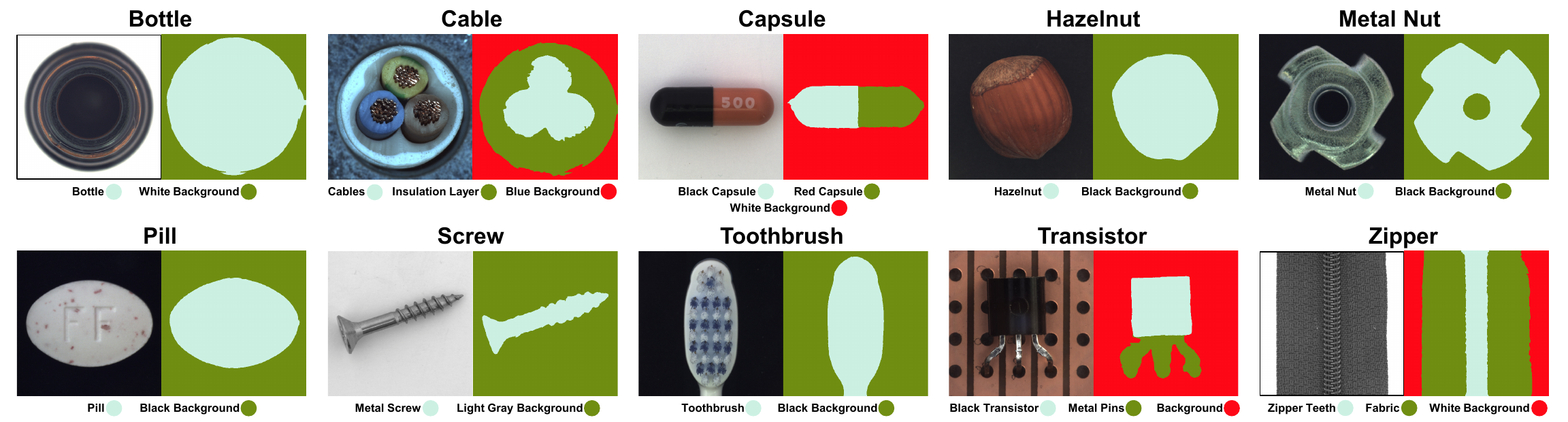}
   \vspace{-0.15cm}
   \caption{Examples of component segmentation on the MVTec-AD dataset. Texture classes are not displayed as they contain only a single component.}
   \label{fig:figs2}
   \vspace{-0.15cm}
\end{figure}

\begin{figure}[t]
  \centering
   \includegraphics[width=0.93\linewidth]{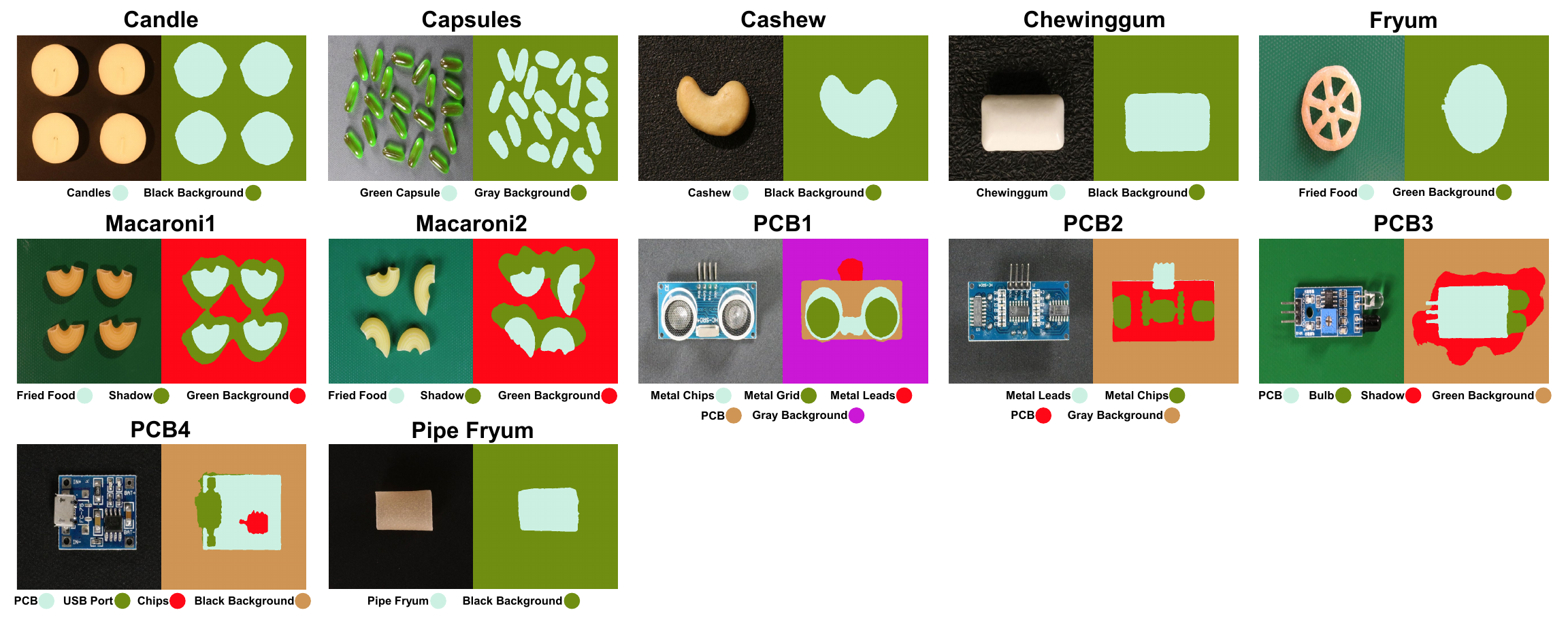}
      \vspace{-0.15cm}
   \caption{Examples of component segmentation on the VisA dataset.}
   \label{fig:figs3}
   \vspace{-0.15cm}
\end{figure}

\begin{figure}[t]
  \centering
   \includegraphics[width=0.93\linewidth]{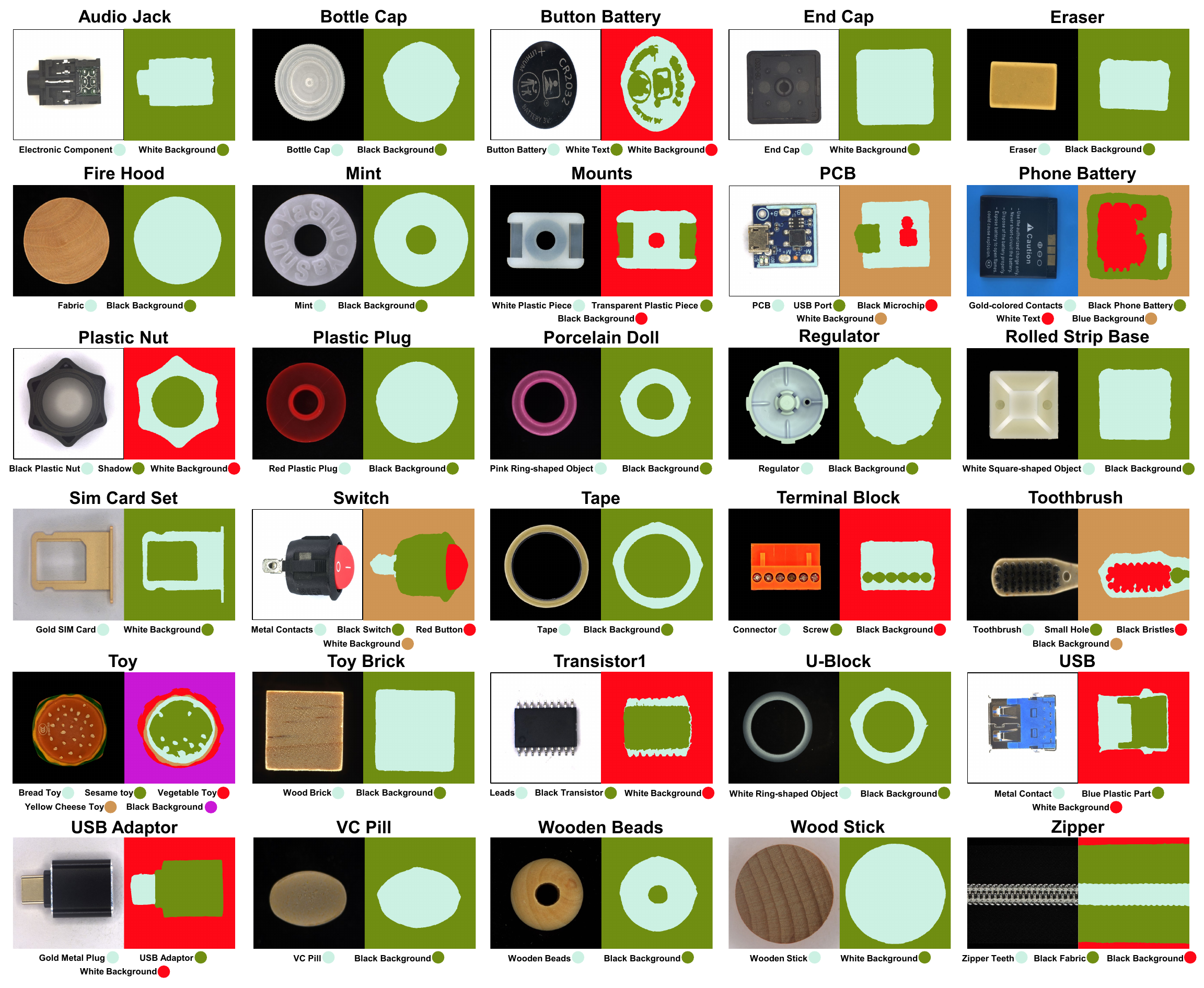}
   \vspace{-0.15cm}
   \caption{Examples of component segmentation on the Real-IAD dataset.}
   \label{fig:figs4}
\end{figure}

\begin{figure}[t]
  \centering
     \vspace{-0.5cm}
   \includegraphics[width=0.85\linewidth]{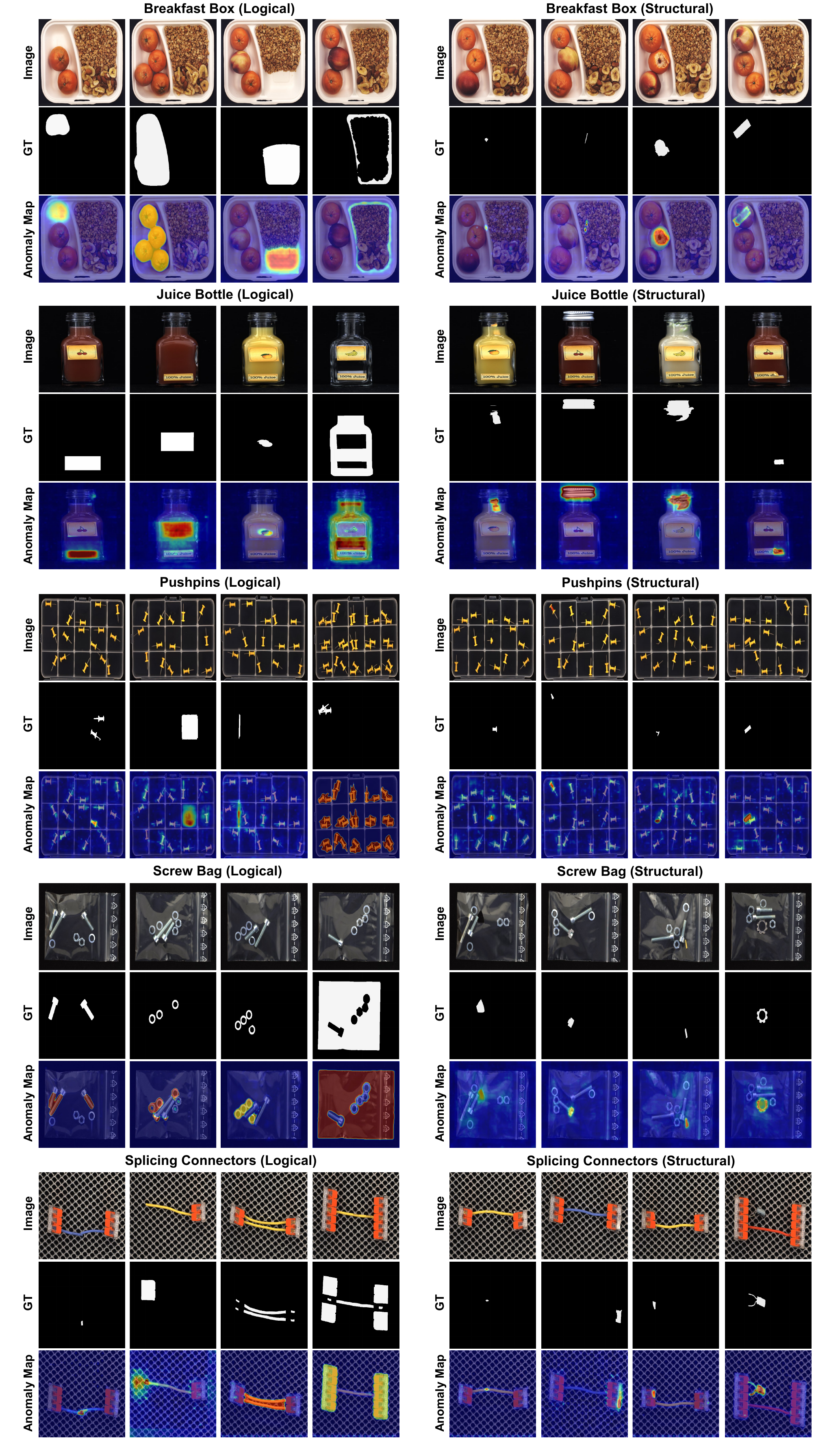}
   \caption{Qualitative results of LogiCo on the MVTec-LOCO dataset.}
   \label{fig:figs5}
\end{figure}

\begin{figure}[t]
  \centering
     \vspace{-0.5cm}
   \includegraphics[width=0.85\linewidth]{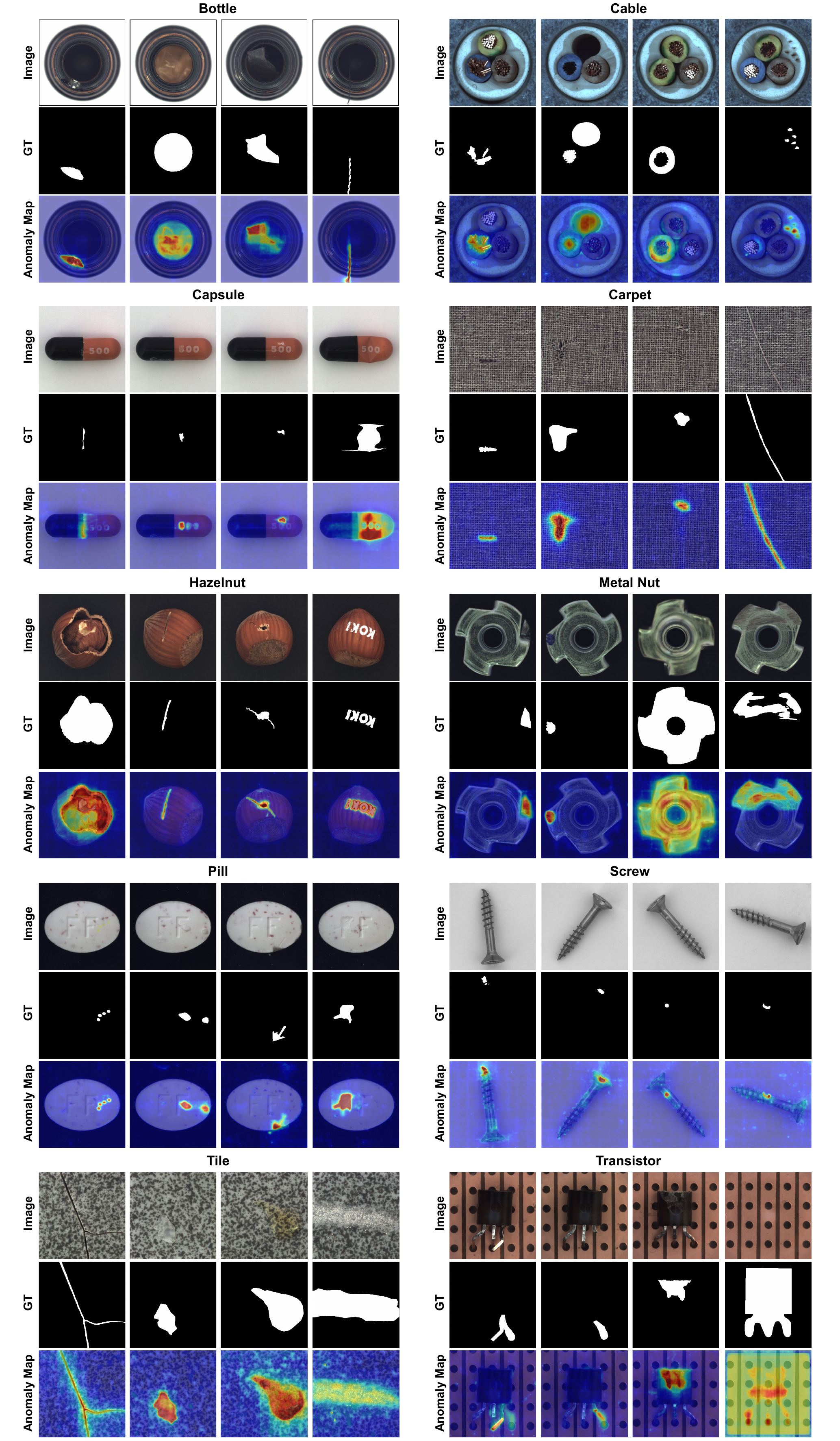}
   \caption{Qualitative results of LogiCo on the MVTec-AD dataset.}
   \label{fig:figs6}
\end{figure}

\begin{figure}[t]
  \centering
     \vspace{-0.5cm}
   \includegraphics[width=0.85\linewidth]{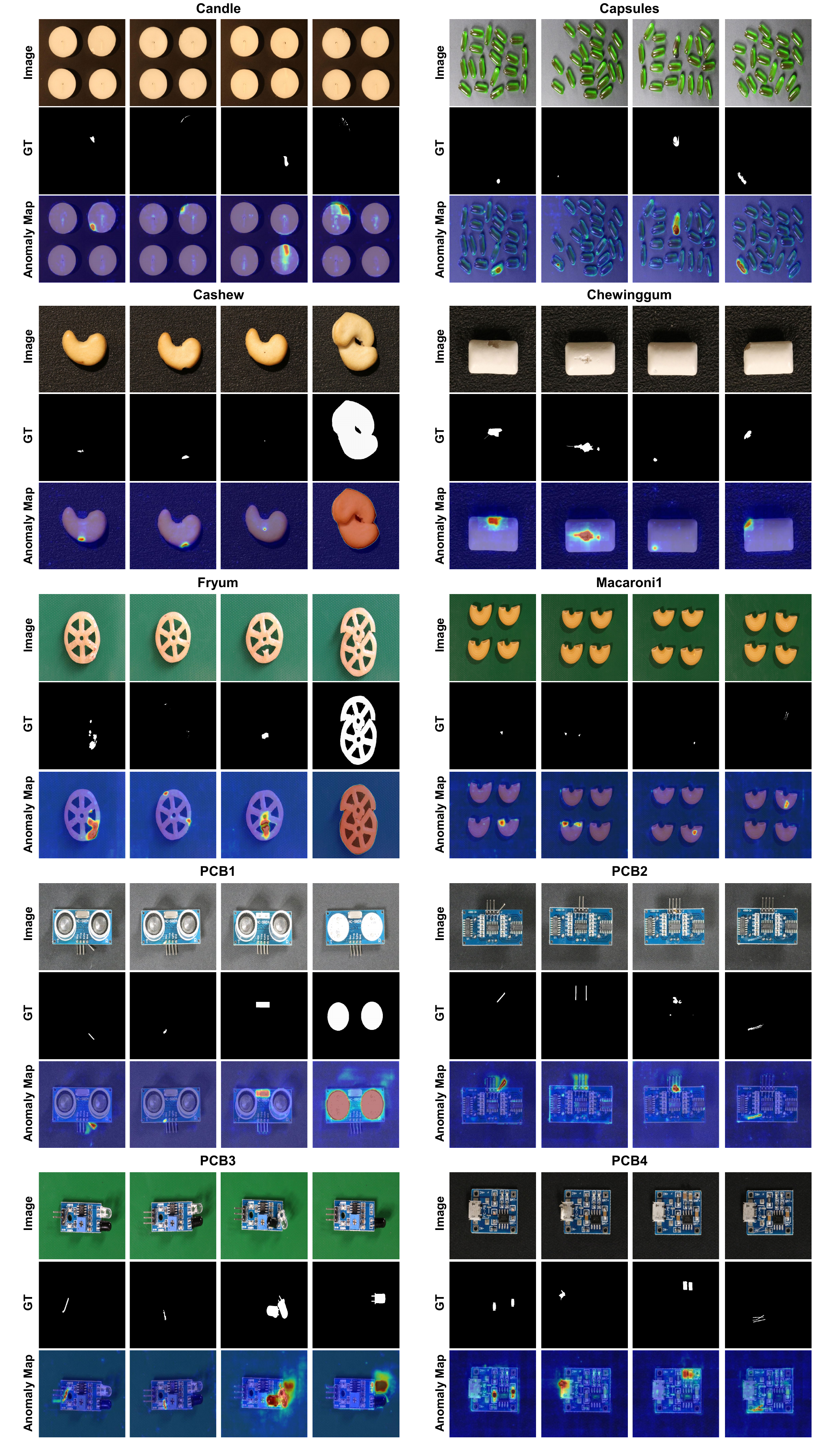}
   \caption{Qualitative results of LogiCo on the VisA dataset.}
   \label{fig:figs7}
\end{figure}

\begin{figure}[t]
  \centering
     \vspace{-0.5cm}
   \includegraphics[width=0.85\linewidth]{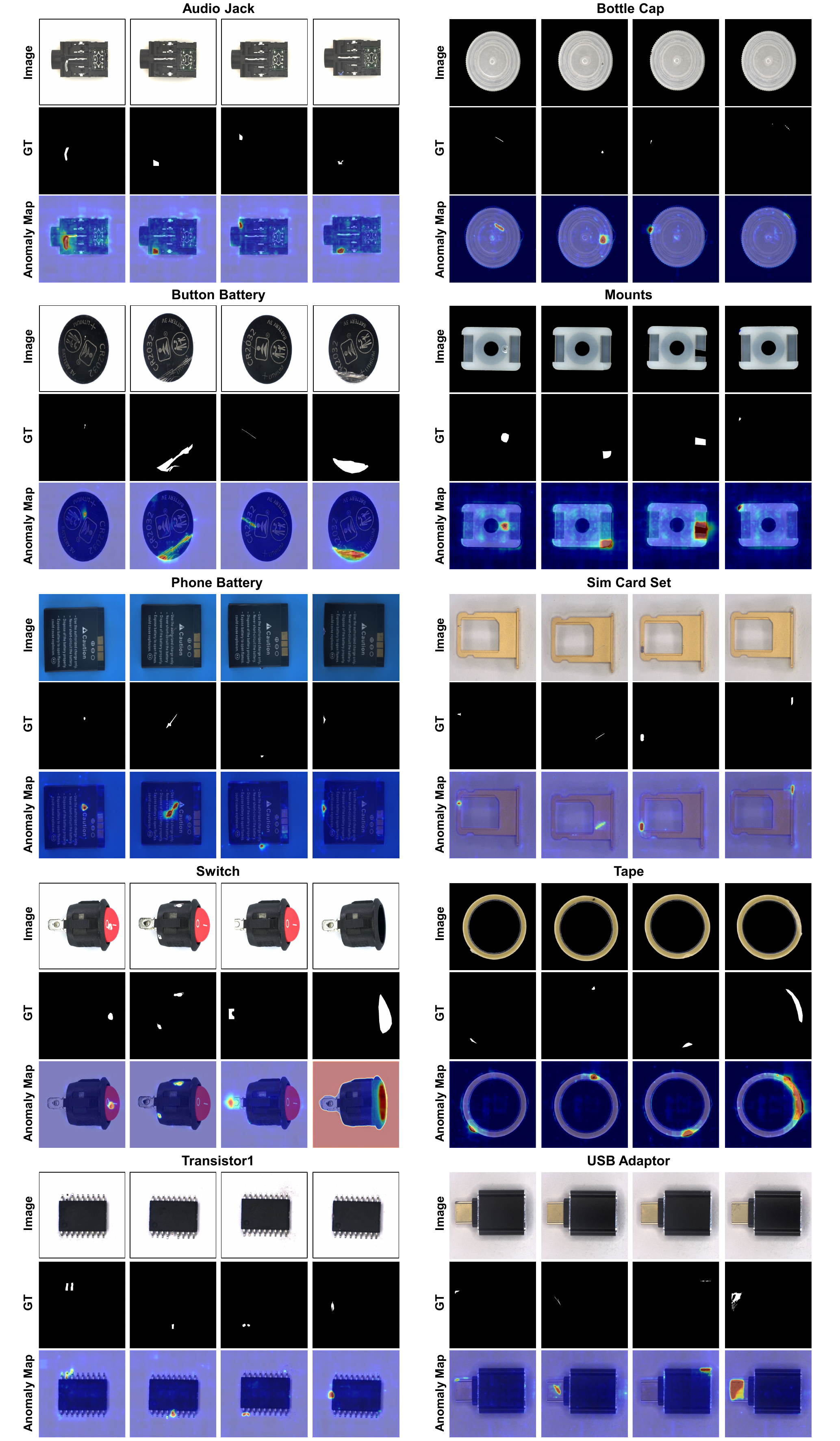}
   \caption{Qualitative results of LogiCo on the Real-IAD dataset.}
   \label{fig:figs8}
\end{figure}

\textbf{Supplementary Qualitative Results and Per-Category Metrics.} Figs. \ref{fig:figs2}–\ref{fig:figs4} respectively display the component segmentation results on the MVTec-AD \cite{bergmann2019mvtec}, VisA \cite{zou2022spot}, and Real-IAD \cite{wang2024real} datasets. For most categories, we only use the image category name and a simple background description as prompts, without requiring complex prompt tuning, which highlights LogiCo's strong generalization and transferability. Figs. \ref{fig:figs5}–\ref{fig:figs8} provide supplementary qualitative results for LogiCo, while Tabs. \ref{tab:tables3}–\ref{tab:tables7} report per-category metrics for LogiCo and other methods. Extensive experiments across four benchmarks \cite{bergmann2022beyond, bergmann2019mvtec, zou2022spot, wang2024real} demonstrate LogiCo's broad applicability across diverse anomaly detection tasks.

\section{Limitations}

LogiCo integrates the SMD module to detect count-related anomalies. However, since SMD lacks fine-grained intra-component information, it may lead to degraded anomaly localization performance. In addition, as LogiCo involves a component segmentation process, it may incur higher computational cost than some efficient structural anomaly detection methods. Nevertheless, LogiCo still achieves fast inference at 17.4 FPS under a resolution of 512 $\times$ 512, which satisfies the real-time detection requirements of real-world application scenarios.

\clearpage

\clearpage

\clearpage

\begin{table}[h]
  \centering
  \renewcommand\arraystretch{1.15}
  \vspace{0.1cm}
  \caption{Comparison of anomaly detection performance between LogiCo and other methods on the MVTec-LOCO dataset, employing Logical-AUC (\%), Structural-AUC (\%), and Mean-AUC (\%) as metrics.}
  \resizebox{\linewidth}{!}{
    \begin{tabular}{ccccccc}
    \toprule
    Category & PatchCore\cite{roth2022towards} & INP-Former\cite{luo2025exploring} & Dinomaly\cite{guo2025dinomaly} & CSAD\cite{Hsieh_2024_BMVC}  & SALAD\cite{fuvcka2025salad} & \cellcolor{gray!15}\textbf{LogiCo} \\
    \midrule
    Breakfast Box & \:\:\:84.12/73.75/78.94\:\:\: & \:\:\:84.08/93.22/88.65\:\:\: & \:\:\:88.19/\underline{97.32}/92.75\:\:\: & \:\:\:\underline{96.91}/91.52/\underline{94.22}\:\:\: & \:\:\:88.59/85.38/86.99\:\:\: & \cellcolor{gray!15}\:\:\:\textbf{99.02}/\textbf{98.65}/\textbf{98.83}\:\:\: \\
    Juice Bottle & 83.10/88.92/86.01 & 87.83/95.02/91.43 & 89.03/\underline{98.92}/93.98 & 91.11/92.96/92.04 & \textbf{99.93}/\textbf{99.58}/\textbf{99.75} & \cellcolor{gray!15}\underline{99.71}/98.14/\underline{98.93} \\
    Pushpins & 64.48/77.49/70.99 & 54.47/76.44/65.45 & 58.19/77.46/67.82 & \underline{87.74}/\underline{97.18}/\underline{92.46} & 86.06/\textbf{98.17}/92.12 & \cellcolor{gray!15}\textbf{99.63}/90.73/\textbf{95.18} \\
    Screw Bag & 54.77/89.52/72.15 & 47.56/83.48/65.52 & 50.39/92.51/71.45 & \textbf{97.42}/92.68/\textbf{95.05} & 88.39/\underline{93.49}/90.94 & \cellcolor{gray!15}\underline{91.18}/\textbf{93.63}/\underline{92.41} \\
    Splicing Connectors & 61.28/78.21/69.75 & 77.09/93.70/85.39 & 82.73/\underline{98.97}/90.85 & 89.95/91.36/90.66 & \textbf{95.70}/96.84/\textbf{96.27} & \cellcolor{gray!15}\underline{93.09}/\textbf{99.02}/\underline{96.06} \\
    \midrule
    \textbf{AVG}   & 69.55/81.58/75.57 & 70.21/88.37/79.29 & 73.71/93.04/83.37 & \underline{92.63}/93.14/92.89 & 91.73/\underline{94.69}/\underline{93.21} & \cellcolor{gray!15}\textbf{96.53}/\textbf{96.03}/\textbf{96.28} \\
    \bottomrule
    \end{tabular}%
    }
  \label{tab:tables3}%
    \vspace{-0.7cm}
  \end{table}%

\begin{table}[h]
  \centering
  \renewcommand\arraystretch{1.15}
  \caption{Comparison of anomaly localization performance between LogiCo and other methods on the MVTec-LOCO dataset, employing Logical-sPRO (\%), Structural-sPRO (\%), and Mean-sPRO (\%) as metrics.}
  \resizebox{\linewidth}{!}{
    \begin{tabular}{ccccccc}
    \toprule
    Category & PatchCore\cite{roth2022towards} & INP-Former\cite{luo2025exploring} & Dinomaly\cite{guo2025dinomaly} & CSAD\cite{Hsieh_2024_BMVC}  & SALAD\cite{fuvcka2025salad} & \cellcolor{gray!15}\textbf{LogiCo} \\
    \midrule
    Breakfast Box & \:\:\:39.55/36.87/38.21\:\:\: & \:\:\:\textbf{52.27}/84.68/\textbf{68.47}\:\:\: & \:\:\:40.37/\underline{91.36}/65.87\:\:\: & \:\:\:43.30/75.25/59.28\:\:\: & \:\:\:\underline{45.04}/75.29/60.17\:\:\: & \cellcolor{gray!15}\:\:\:42.72/\textbf{92.58}/\underline{67.65}\:\:\: \\
    Juice Bottle & 33.02/45.75/39.38 & 60.40/74.72/67.56 & 72.54/\textbf{88.97}/80.76 & \textbf{80.44}/86.91/\textbf{83.68} & 70.14/78.43/74.29 & \cellcolor{gray!15}\underline{77.41}/\underline{87.93}/\underline{82.67} \\
    Pushpins & 27.51/17.57/22.54 & 43.40/63.79/53.59 & 57.13/\underline{69.97}/63.55 & \textbf{96.49}/61.67/\underline{79.08} & 58.11/49.65/53.88 & \cellcolor{gray!15}\underline{93.70}/\textbf{71.42}/\textbf{82.56} \\
    Screw Bag & 40.96/50.22/45.59 & 40.74/\underline{73.66}/57.20 & 32.13/\textbf{75.90}/54.02 & 43.14/73.55/58.35 & \underline{52.60}/70.81/\underline{61.71} & \cellcolor{gray!15}\textbf{64.24}/71.50/\textbf{67.87} \\
    Splicing Connectors & 29.72/22.85/26.29 & 51.34/84.94/\underline{68.14} & 26.98/\textbf{92.49}/59.74 & \textbf{64.33}/80.48/\textbf{72.41} & \underline{61.06}/71.90/66.48 & \cellcolor{gray!15}40.10/\underline{87.38}/63.74 \\
    \midrule
    \textbf{AVG}   & 34.15/34.65/34.40 & 49.63/76.36/62.99 & 45.83/\textbf{83.74}/64.79 & \textbf{65.54}/75.57/\underline{70.56} & 57.39/69.22/63.31 & \cellcolor{gray!15}\underline{63.63}/\underline{82.16}/\textbf{72.90} \\
    \bottomrule
    \end{tabular}%
    }
  \label{tab:tables4}%
  \vspace{-0.7cm}
\end{table}%

\begin{table}[h]
\centering
  \renewcommand\arraystretch{1.15}
  \caption{Comparison of anomaly detection and localization performance between LogiCo and other methods on the MVTec-AD dataset, employing I-AUC (\%), P-AUC (\%), P-AP (\%), and PRO (\%) as metrics.}
  \resizebox{\linewidth}{!}{
    \begin{tabular}{ccccccc}
    \toprule
    Category & PatchCore\cite{roth2022towards} & INP-Former\cite{luo2025exploring} & Dinomaly\cite{guo2025dinomaly} & CSAD\cite{Hsieh_2024_BMVC}  & SALAD\cite{fuvcka2025salad} & \cellcolor{gray!15}\textbf{LogiCo} \\
    \midrule
    Bottle & \:\:\textbf{100.0}/98.97/82.87/\underline{95.94}\:\: & \:\:\underline{99.84}/\underline{99.15}/\underline{85.85}/94.97\:\: & \:\:\textbf{100.0}/\textbf{99.24}/\textbf{86.25}/\textbf{97.47}\:\: & \:\:99.37/98.66/80.97/92.77\:\: & \:\:99.60/97.29/60.65/87.42\:\: & \cellcolor{gray!15}\:\:\textbf{100.0}/98.99/84.01/94.74\:\: \\
    Cable &    \underline{99.66}/\underline{98.90}/74.20/\underline{94.77} & 99.14/97.55/\underline{75.61}/91.91 & \textbf{99.79}/\textbf{98.91}/\textbf{75.88}/\textbf{95.62} & 94.96/98.57/71.65/90.05 & 96.61/98.02/65.60/85.09 & \cellcolor{gray!15}99.36/97.78/71.73/92.91 \\
    Capsule & 98.84/99.03/51.79/95.16 & 98.17/98.46/\underline{60.96}/\textbf{97.19} & \underline{98.96}/\underline{99.10}/60.68/95.73 & 94.89/98.26/43.64/89.83 & 96.42/98.00/52.15/83.53 & \cellcolor{gray!15}\textbf{99.40}/\textbf{99.29}/\textbf{63.19}/\underline{96.80}\\
    Carpet & 98.61/99.06/70.14/94.94 & \textbf{100.0}/\underline{99.54}/\textbf{79.54}/\underline{98.18} & \textbf{100.0}/\textbf{99.64}/\underline{78.10}/\textbf{98.79} & 92.70/98.13/46.48/93.23 & 99.04/96.41/66.28/89.88 & \cellcolor{gray!15}99.88/99.31/69.77/97.45 \\
    Grid  & 98.66/98.89/49.05/94.21 & \textbf{100.0}/\underline{99.50}/\textbf{60.96}/\underline{97.26} & \textbf{100.0}/\textbf{99.54}/\underline{56.51}/\textbf{97.65} & 99.08/98.10/37.37/91.07 & 98.58/97.81/33.40/92.37 & \cellcolor{gray!15}\textbf{100.0}/98.65/48.02/94.83 \\
    Hazelnut & \textbf{100.0}/99.00/68.52/96.18 & \textbf{100.0}/99.62/81.31/\underline{97.56} & \textbf{100.0}/\underline{99.64}/\underline{81.56}/\textbf{97.77} & 99.82/98.82/63.27/92.00 & 99.54/97.93/54.11/93.16 & \cellcolor{gray!15}\textbf{100.0}/\textbf{99.67}/\textbf{90.11}/95.25 \\
    Leather & \textbf{100.0}/99.45/56.08/97.48 & \textbf{100.0}/\textbf{99.64}/\textbf{61.72}/\textbf{99.19} & \textbf{100.0}/\underline{99.59}/\underline{58.24}/97.20 & 99.69/99.42/48.77/93.55 & \textbf{100.0}/98.47/58.21/88.13 & \cellcolor{gray!15}\textbf{100.0}/99.40/54.08/\underline{98.89}\\
    Metal Nut & 99.90/\underline{98.81}/\underline{90.45}/\underline{95.75} & \textbf{100.0}/97.71/76.95/95.69 & \textbf{100.0}/98.37/82.57/\textbf{96.65} & 99.02/98.44/85.73/93.16 & 97.07/94.61/54.49/75.17 & \cellcolor{gray!15}\textbf{100.0}/\textbf{99.37}/\textbf{94.21}/94.40 \\
    Pill  & 96.84/98.05/\textbf{79.80}/95.20 & 99.20/96.35/58.01/97.05 & \textbf{99.51}/\underline{98.54}/75.54/\textbf{98.38} & 91.23/95.20/53.29/80.44 & 95.11/96.51/50.82/73.59 & \cellcolor{gray!15}\underline{99.40}/\textbf{98.88}/\underline{78.76}/\underline{97.30} \\
    Screw & 97.01/\underline{99.43}/47.21/\textbf{96.72} & \underline{97.96}/99.30/\textbf{60.51}/\underline{96.56} & \textbf{98.32}/\textbf{99.64}/\underline{59.15}/96.33 & 79.30/96.66/15.45/86.26 & 83.91/98.12/25.82/88.70 & \cellcolor{gray!15}97.32/99.21/53.99/93.89 \\
    Tile  & \textbf{100.0}/96.86/63.89/90.54 & \textbf{100.0}/\underline{98.07}/74.47/\underline{94.40} & \textbf{100.0}/\textbf{98.48}/\underline{78.56}/94.27 & 93.97/96.20/71.94/43.53 & 99.82/96.87/76.42/72.88 & \cellcolor{gray!15}\textbf{100.0}/97.77/\textbf{81.03}/\textbf{94.48} \\
    Toothbrush & \textbf{100.0}/99.01/63.51/91.56 & \textbf{100.0}/99.38/61.32/\underline{95.83} & \textbf{100.0}/\textbf{99.51}/\underline{67.22}/\textbf{97.08} & 98.06/97.65/40.71/78.17 & 98.33/96.60/25.60/58.92 & \cellcolor{gray!15}99.17/\underline{99.47}/\textbf{74.82}/92.97 \\
    Transistor & 99.62/96.32/65.33/90.16 & 98.67/94.96/62.40/84.56 & 99.29/94.63/60.79/76.55 & \textbf{100.0}/\textbf{98.52}/\textbf{83.09}/\textbf{96.27} & 99.29/\underline{97.65}/\underline{70.59}/92.06 & \cellcolor{gray!15}\underline{99.71}/97.55/61.53/\underline{92.98} \\
    Wood  & 98.86/95.33/56.41/91.23 & 99.12/\textbf{98.36}/\underline{73.80}/\textbf{96.54} & 98.51/\underline{97.59}/\textbf{75.83}/\underline{96.45} & \underline{99.39}/94.54/47.52/86.03 & 98.42/94.92/56.03/77.50 & \cellcolor{gray!15}\textbf{99.82}/95.88/60.75/93.12\\
    Zipper & 99.16/\underline{99.04}/67.88/\underline{96.65} & \underline{99.82}/98.51/\underline{70.16}/95.69 & \textbf{100.0}/\textbf{99.30}/\textbf{70.33}/\textbf{97.02} & 95.41/97.91/55.26/92.15 & 95.91/94.00/38.76/57.82 & \cellcolor{gray!15}99.76/98.94/66.88/94.98\\
    \midrule
    \textbf{AVG}   & 99.14/98.41/65.81/94.43 & 99.46/98.41/69.57/\underline{95.51} & \textbf{99.63}/\textbf{98.78}/\textbf{71.15}/\textbf{95.53} & 95.79/97.67/56.34/86.57 & 97.18/96.88/52.60/81.08 & \cellcolor{gray!15}\underline{99.59}/\underline{98.68}/\underline{70.19}/95.00\\
    \bottomrule
    \end{tabular}%
    }
  \label{tab:tables5}%
\end{table}%

\clearpage

\begin{table}[h]
   \centering
   \renewcommand\arraystretch{1.15}
    \caption{Comparison of anomaly detection and localization performance between LogiCo and other methods on the VisA dataset, employing I-AUC (\%), P-AUC (\%), P-AP (\%), and PRO (\%) as metrics.}
      \resizebox{\linewidth}{!}{
    \begin{tabular}{ccccccc}
    \toprule
    Category & PatchCore\cite{roth2022towards} & INP-Former\cite{luo2025exploring} & Dinomaly\cite{guo2025dinomaly} & CSAD\cite{Hsieh_2024_BMVC}  & SALAD\cite{fuvcka2025salad} & \cellcolor{gray!15}\textbf{LogiCo} \\
    \midrule
    Candle & \:\:\underline{98.61}/99.24/33.12/94.35\:\: & \:\:\textbf{98.76}/99.63/43.33/\underline{97.05}\:\: & \:\:98.48/\underline{99.66}/\textbf{51.83}/94.51\:\: & \:\:94.71/97.81/16.36/91.89\:\: & \:\:97.22/92.43/25.53/63.68\:\: & \cellcolor{gray!15}\:\:98.24/\textbf{99.68}/\underline{47.84}/\textbf{97.84}\:\: \\
    Capsules & 79.35/99.18/\textbf{70.88}/89.87 & 97.87/98.36/61.44/93.83 & \textbf{99.28}/\textbf{99.69}/64.67/\underline{95.59} & 90.90/97.86/27.94/73.46 & 92.28/93.43/28.39/67.90 & \cellcolor{gray!15}\underline{99.25}/\underline{99.49}/\underline{69.53}/\textbf{96.50} \\
    Cashew & 97.46/98.62/62.65/93.02 & 97.86/96.16/56.53/93.82 & \textbf{98.90}/\underline{99.34}/\underline{70.85}/\textbf{97.93} & 91.50/98.58/64.13/76.72 & 97.82/99.22/59.78/82.15 & \cellcolor{gray!15}\underline{98.78}/\textbf{99.85}/\textbf{97.63}/\underline{96.14} \\
    Chewinggum & 99.12/98.38/47.03/83.67 & 99.26/\textbf{99.66}/\underline{75.72}/\textbf{94.27} & 99.24/\underline{99.58}/65.91/\underline{94.18} & 95.70/98.15/18.53/69.74 & \textbf{99.72}/98.50/25.07/72.48 & \cellcolor{gray!15}\underline{99.46}/99.57/\textbf{82.04}/91.83 \\
    Fryum & 95.38/93.87/45.33/83.66 & \underline{99.62}/96.66/41.78/\underline{94.48} & 99.34/\textbf{97.37}/\underline{54.38}/\textbf{95.94} & \textbf{99.66}/92.32/29.63/70.15 & 98.48/96.11/36.15/88.87 & \cellcolor{gray!15}98.80/\underline{97.06}/\textbf{54.85}/90.39 \\
    Macaroni1 & 97.47/99.40/31.95/95.96 & 97.46/99.82/31.74/\underline{97.91} & 98.59/\textbf{99.88}/\textbf{43.97}/96.77 & 85.35/97.03/13.35/79.44 & \textbf{99.31}/99.18/25.50/93.54 & \cellcolor{gray!15}\underline{98.67}/\textbf{99.88}/\underline{37.87}/\textbf{98.52} \\
    Macaroni2 & 79.45/98.49/15.36/94.83 & \underline{95.61}/99.76/24.98/\underline{97.72} & \textbf{98.09}/\textbf{99.88}/\textbf{36.25}/93.47 & 80.53/97.05/14.71/82.94 & 90.52/96.88/22.89/83.02 & \cellcolor{gray!15}94.49/\underline{99.84}/\underline{29.88}/\textbf{98.20} \\
    PCB1  & 98.34/99.64/86.92/\underline{94.47} & 97.88/99.25/85.37/91.52 & \textbf{98.98}/\underline{99.72}/\textbf{88.14}/\textbf{95.08} & 93.80/99.45/59.54/84.70 & \underline{98.94}/99.31/51.51/75.88 & \cellcolor{gray!15}97.50/\textbf{99.84}/\underline{88.08}/92.77 \\
    PCB2  & 96.97/\underline{98.75}/27.17/\underline{87.16} & 97.87/\textbf{98.77}/\underline{32.16}/81.19 & 98.12/98.67/\textbf{50.86}/\textbf{89.19} & 93.21/95.96/20.28/65.57 & \textbf{99.30}/98.07/29.94/76.94 & \cellcolor{gray!15}\underline{98.22}/98.60/25.94/85.36 \\
    PCB3  & 97.81/98.91/40.99/90.49 & \underline{99.30}/99.15/40.07/\underline{94.87} & \textbf{99.43}/\underline{99.32}/\underline{47.41}/\textbf{95.34} & 85.17/97.61/25.86/62.77 & 97.01/99.10/22.36/90.17 & \cellcolor{gray!15}98.98/\textbf{99.34}/\textbf{47.77}/89.75 \\
    PCB4  & \underline{99.58}/98.00/\underline{48.55}/88.86 & \textbf{99.80}/98.23/44.66/\underline{92.52} & 99.53/\textbf{99.10}/\textbf{50.43}/\textbf{95.45} & 85.66/94.69/30.10/56.12 & 99.20/\underline{98.33}/27.45/64.90 & \cellcolor{gray!15}99.56/98.14/45.27/92.28 \\
    Pipe Fryum & \underline{99.74}/99.15/65.98/\underline{95.56} & 99.26/99.18/52.88/\textbf{97.96} & 99.40/\underline{99.43}/\underline{67.72}/94.28 & 90.56/97.91/52.29/71.02 & 99.38/98.52/37.45/49.20 & \cellcolor{gray!15}\textbf{99.78}/\textbf{99.86}/\textbf{94.09}/95.12 \\
    \midrule
    \textbf{AVG} & 94.94/98.47/47.99/90.99 & 98.38/98.72/49.22/\underline{93.93} & \textbf{98.95}/\textbf{99.30}/\underline{57.70}/\textbf{94.81} & 90.56/97.04/31.06/73.71 & 97.43/97.42/32.67/75.73 & \cellcolor{gray!15}\underline{98.48}/\underline{99.26}/\textbf{60.07}/93.73 \\
    \bottomrule
    \end{tabular}%
    }
  \label{tab:tables6}%
    \vspace{-0.7cm}
\end{table}%

\begin{table}[h]
   \centering
   \renewcommand\arraystretch{1.15}
  \caption{Comparison of anomaly detection and localization performance between LogiCo and other methods on the Real-IAD dataset under the single-view setting, employing I-AUC (\%), P-AUC (\%), P-AP (\%), and PRO (\%) as metrics.}
   \resizebox{\linewidth}{!}{
     \begin{tabular}{ccccccc}
    \toprule
    Category & PatchCore\cite{roth2022towards} & INP-Former\cite{luo2025exploring} & Dinomaly\cite{guo2025dinomaly} & CSAD\cite{Hsieh_2024_BMVC}  & SALAD\cite{fuvcka2025salad} & \cellcolor{gray!15}\textbf{LogiCo} \\
    \midrule
    Audio Jack & \:\:84.45/98.93/49.52/91.73\:\: & \:\:\underline{93.43}/99.72/\underline{58.15}/94.71\:\: & \:\:\textbf{93.97}/\textbf{99.88}/55.53/\textbf{98.75}\:\: & \:\:84.94/96.26/29.24/72.11\:\: & \:\:86.16/95.62/26.43/71.70\:\: &  \cellcolor{gray!15}\:\:92.09/\underline{99.73}/\textbf{62.16}/\underline{96.85}\:\: \\
    Bottle Cap & 95.65/99.41/21.46/\textbf{95.44} & \underline{96.06}/99.37/31.30/94.30 & 95.70/\textbf{99.71}/30.93/95.04 & 83.67/94.64/27.43/73.10 & 90.51/94.14/\underline{35.47}/74.12 & \cellcolor{gray!15}\textbf{96.71}/\underline{99.57}/\textbf{45.40}/\underline{95.25} \\
    Button Battery & 87.14/99.43/69.65/91.59 & \underline{90.47}/99.23/\textbf{71.44}/92.32 & 89.58/\textbf{99.70}/\underline{70.25}/\textbf{96.06} & 82.33/96.82/27.96/69.80 & 88.52/94.51/30.09/72.56 & \cellcolor{gray!15}\textbf{90.81}/\underline{99.44}/46.69/\underline{94.31} \\
    End Cap & 84.40/97.04/9.36/89.89 & 82.44/\underline{98.79}/16.24/92.30 & \textbf{85.51}/\textbf{99.53}/14.06/\textbf{97.70} & 62.03/91.83/\textbf{21.79}/77.33 & 69.63/89.62/20.45/59.87 & \cellcolor{gray!15}\underline{84.91}/98.41/\underline{20.77}/\underline{95.97} \\
    Eraser & 91.87/99.37/36.11/93.50 & \underline{95.68}/\textbf{99.89}/\textbf{54.05}/95.08 & 93.18/\underline{99.87}/49.20/\underline{97.72} & 84.89/96.77/36.44/67.36 & 83.33/98.79/23.93/83.03 & \cellcolor{gray!15}\textbf{96.35}/99.72/\underline{51.94}/\textbf{98.20} \\
    Fire Hood & 94.14/98.95/37.65/93.61 & 93.89/\underline{99.82}/55.83/\underline{97.18} & \underline{94.66}/\textbf{99.85}/\underline{60.24}/\textbf{98.02} & 91.52/98.23/35.69/88.13 & 85.56/97.22/20.73/78.33 & \cellcolor{gray!15}\textbf{95.78}/99.79/\textbf{65.27}/95.87 \\
    Mint  & 67.86/96.71/30.37/67.81 & \textbf{80.51}/\underline{99.16}/\underline{36.52}/85.34 & \underline{79.38}/\textbf{99.61}/35.28/\textbf{87.88} & 69.50/95.04/18.22/63.84 & 67.45/96.01/19.25/65.18 & \cellcolor{gray!15}73.54/98.50/\textbf{36.58}/\underline{85.72} \\
    Mounts & 95.08/99.62/38.26/98.16 & 96.75/99.86/\underline{51.69}/\textbf{99.35} & \underline{97.27}/\underline{99.87}/49.98/\textbf{99.35} & 81.33/94.89/23.85/79.58 & 89.71/97.88/28.50/72.17 & \cellcolor{gray!15}\textbf{97.34}/\textbf{99.88}/\textbf{66.98}/98.52 \\
    PCB   & 93.84/99.47/53.10/\underline{95.92} & 92.09/99.13/48.61/\textbf{96.12} & \textbf{96.26}/\textbf{99.88}/\underline{57.20}/95.59 & 82.06/97.16/37.33/78.80 & \underline{95.90}/98.19/29.20/85.42 & \cellcolor{gray!15}93.89/\underline{99.60}/\textbf{68.54}/92.90 \\
    Phone Battery & 92.27/98.74/39.87/92.16 & \underline{95.60}/\underline{99.90}/\textbf{63.47}/\textbf{98.75} & 95.53/\textbf{99.91}/60.16/96.89 & 92.01/98.11/26.09/88.07 & 88.77/97.00/23.63/77.41 & \cellcolor{gray!15}\textbf{97.47}/99.82/\underline{63.32}/\underline{97.64} \\
    Plastic Nut & 88.81/99.24/30.84/94.74 & 92.26/\underline{99.81}/\underline{46.37}/\underline{97.27} & \textbf{95.30}/\textbf{99.86}/45.41/\textbf{97.48} & 79.17/95.27/23.42/77.25 & 85.21/94.93/24.73/71.47 & \cellcolor{gray!15}\underline{94.60}/99.62/\textbf{55.62}/94.68 \\
    Plastic Plug & 95.85/99.50/\underline{39.67}/97.05 & \underline{96.10}/\underline{99.71}/36.45/\textbf{99.09} & \textbf{96.21}/\textbf{99.83}/36.26/97.26 & 89.88/98.81/27.59/94.63 & 92.39/98.36/28.27/89.64 & \cellcolor{gray!15}95.57/99.53/\textbf{46.24}/\underline{97.83} \\
    Porcelain Doll & 88.81/98.84/20.62/93.71 & \underline{93.36}/\underline{99.62}/\underline{29.11}/\textbf{98.97} & \textbf{95.38}/\textbf{99.84}/28.60/92.31 & 80.81/95.28/27.36/67.09 & 88.05/97.62/23.69/80.59 & \cellcolor{gray!15}92.70/99.07/\textbf{34.89}/\underline{96.96} \\
    Regulator & \underline{95.86}/99.90/48.16/\underline{99.37} & 95.71/99.87/57.96/97.29 & 94.04/\textbf{99.97}/\underline{60.86}/\textbf{99.67} & 80.11/85.53/25.12/62.04 & 93.78/92.42/36.50/82.37 & \cellcolor{gray!15}\textbf{97.14}/\underline{99.94}/\textbf{71.01}/98.68 \\
    Rolled Strip Base & 99.42/99.63/29.94/98.40 & \textbf{99.57}/\textbf{99.88}/\underline{48.40}/\textbf{99.55} & \underline{99.52}/\underline{99.84}/39.66/\underline{99.07} & 99.28/98.66/24.91/95.19 & 99.38/95.39/37.62/90.72 &  \cellcolor{gray!15}99.50/99.80/\textbf{57.87}/98.84 \\
    Sim Card Set & 98.36/99.61/55.15/97.25 & \underline{98.75}/\underline{99.84}/\underline{67.16}/\textbf{98.75} & \textbf{99.05}/\textbf{99.87}/65.90/\underline{98.51} & 97.97/99.57/48.76/93.83 & 98.18/97.69/37.01/83.70 & \cellcolor{gray!15}97.83/99.77/\textbf{71.11}/97.37 \\
    Switch & 97.08/\textbf{99.35}/\underline{60.64}/95.50 & \underline{97.40}/97.50/56.05/\textbf{96.55} & \textbf{98.92}/98.86/56.48/\underline{95.82} & 80.92/98.22/57.07/78.22 & 96.07/97.86/40.89/78.86 & \cellcolor{gray!15}96.53/\underline{99.34}/\textbf{60.65}/94.43 \\
    Tape  & 98.92/99.78/41.51/99.06 & 99.26/99.91/49.07/\underline{99.60} & \textbf{99.57}/\textbf{99.93}/\underline{51.03}/99.16 & 66.14/95.02/15.45/85.63 & 97.97/97.63/23.54/75.21 & \cellcolor{gray!15}\underline{99.49}/\underline{99.92}/\textbf{61.59}/\textbf{99.62} \\
    Terminal Block & 98.45/99.79/37.13/98.69 & \underline{98.82}/99.61/\underline{56.05}/\textbf{98.91} & \textbf{98.97}/\underline{99.89}/51.45/95.50 & 72.32/97.32/19.64/85.29 & 98.59/98.68/28.38/85.63 & \cellcolor{gray!15}98.29/\textbf{99.93}/\textbf{68.44}/\underline{98.90} \\
    Toothbrush & 89.38/99.22/\textbf{56.98}/93.92 & 91.82/\underline{99.40}/42.61/\textbf{97.27} & 91.22/\textbf{99.42}/42.61/\underline{96.76} & \textbf{94.73}/97.88/29.58/88.03 & 89.09/96.45/32.17/84.99 & \cellcolor{gray!15}\underline{92.96}/99.08/\underline{49.22}/94.40 \\
    Toy   & \textbf{92.41}/\textbf{97.99}/16.59/\textbf{92.91} & 86.94/93.74/18.43/85.24 & 89.93/91.10/21.49/84.22 & 78.55/94.51/24.60/72.61 & 86.61/90.09/\textbf{33.68}/74.07 & \cellcolor{gray!15}\underline{92.32}/\underline{97.34}/\underline{26.84}/\underline{90.78} \\
    Toy Brick & 84.23/98.15/39.82/86.77 & \underline{85.07}/\textbf{99.29}/\textbf{55.20}/\textbf{92.79} & 83.34/\underline{99.04}/\underline{54.26}/87.12 & 81.66/97.74/32.90/80.74 & 75.19/93.24/30.92/65.22 & \cellcolor{gray!15}\textbf{88.26}/98.66/50.53/\underline{88.81} \\
    Transistor1 & \underline{96.34}/\underline{99.48}/38.43/\underline{97.54} & 90.76/98.30/37.60/92.55 & 94.88/\textbf{99.58}/\textbf{44.46}/\textbf{97.72} & 86.93/97.80/22.38/79.62 & 95.71/97.94/35.39/88.66 & \cellcolor{gray!15}\textbf{96.57}/99.35/\underline{40.13}/95.63 \\
    U-block & 93.50/99.67/27.07/\underline{97.26} & 91.30/\underline{99.79}/\underline{44.71}/95.96 & \textbf{95.02}/\textbf{99.92}/41.54/\textbf{99.01} & 81.95/97.44/32.39/72.49 & 88.38/97.96/33.10/81.31 & \cellcolor{gray!15}\underline{94.67}/99.67/\textbf{54.94}/94.71 \\
    USB   & \underline{97.06}/99.72/44.06/\textbf{98.06} & 95.19/99.70/46.63/95.98 & \textbf{97.20}/\textbf{99.88}/\underline{47.91}/96.38 & 83.89/97.96/25.69/84.44 & 85.80/98.66/24.67/76.75 & \cellcolor{gray!15}95.22/\underline{99.73}/\textbf{51.89}/\underline{97.90} \\
    USB Adaptor & 89.20/98.52/20.95/90.23 & \underline{92.18}/\underline{99.74}/27.66/\textbf{97.14} & \textbf{92.63}/\textbf{99.82}/25.34/\underline{95.86} & 88.41/97.68/\underline{29.90}/86.24 & 86.03/94.52/25.06/65.93 & \cellcolor{gray!15}90.37/99.65/\textbf{37.24}/95.04 \\
    VC Pill & 92.96/98.45/62.33/86.89 & \textbf{95.18}/\textbf{99.70}/\textbf{77.24}/\textbf{97.23} & \underline{94.83}/\underline{99.51}/\underline{76.76}/\underline{95.02} & 85.62/95.42/37.83/69.39 & 94.16/98.76/47.29/83.56 & \cellcolor{gray!15}94.03/99.23/64.78/90.78 \\
    Wooden Beads & 84.88/97.19/31.47/81.54 & \textbf{90.40}/99.18/47.28/\underline{90.16} & 86.43/\underline{99.37}/\underline{48.49}/90.10 & 78.39/95.76/21.93/66.97 & 84.03/94.31/26.97/71.95 & \cellcolor{gray!15}\underline{90.27}/\textbf{99.38}/\textbf{53.83}/\textbf{92.43} \\
    Wood Stick & 82.12/96.33/36.76/80.29 & \textbf{91.35}/\underline{99.47}/\textbf{69.51}/\underline{95.55} & 89.22/\textbf{99.74}/\underline{62.51}/\textbf{96.54} & 87.88/97.55/46.19/87.20 & 86.37/94.60/29.79/70.87 & \cellcolor{gray!15}\underline{89.59}/98.88/54.43/94.33 \\
    Zipper & \textbf{99.74}/\underline{99.15}/\textbf{65.98}/95.56 & 98.41/98.96/50.91/\underline{96.24} & 99.32/\textbf{99.42}/\underline{60.14}/95.41 & 92.48/96.52/37.19/80.60 & 98.37/98.89/48.35/95.28 & \cellcolor{gray!15}\underline{99.49}/98.94/59.14/\textbf{96.34} \\
    \midrule
    \textbf{AVG} & 91.67/98.91/39.65/92.82 & 93.23/99.26/\underline{48.39}/\underline{95.58} & \underline{93.73}/\textbf{99.42}/48.13/\textbf{95.73} & 83.38/96.32/29.80/78.85 & 88.83/96.17/30.19/77.89 & \cellcolor{gray!15}\textbf{93.81}/\underline{99.38}/\textbf{53.27}/95.32 \\
    \bottomrule
    \end{tabular}%
    }
  \label{tab:tables7}%
\end{table}%

\end{document}